\documentclass[11pt]{article}
\usepackage{coling2018}
\usepackage{times}
\usepackage{url}
\pdfoutput=1
\usepackage{latexsym}
\usepackage{booktabs}
\usepackage{multirow}
\usepackage{pgfplotstable}
\usepackage{filecontents}
\usepackage{array}
\usepackage{xcolor,colortbl}
\usepackage{ctable}
\usepackage{float}
\usepackage{subcaption}
\usepackage{array}
\usepackage{nicefrac}
\usepackage{amsmath}
\usepackage[bottom]{footmisc}
\usepackage[margin=0.1cm]{caption}
\newcolumntype{?}{!{\vrule width 1.5pt}}
\newcommand{\BW}{\cellcolor{black} \color{white} \bfseries}
\newcommand{\G}{\cellcolor{black!25} }
\newcommand{\GG}{\cellcolor{black!40}  \bfseries}
\newrobustcmd{\B}{\bfseries}
\newcommand{\norm}[1]{\left\lVert#1\right\rVert}

\title{Evaluation of Unsupervised Compositional Representations}

\author{Hanan Aldarmaki \\
  The George Washington University \\
 {\tt aldarmaki@gwu.edu} \\\And
 Mona Diab\\
The George Washington University  \\
{\tt mtdiab@gwu.edu} \\}
\date{}

\begin{document}
\maketitle
\begin{abstract}
We evaluated various compositional models, from bag-of-words representations to compositional RNN-based models, on several extrinsic supervised and unsupervised evaluation benchmarks. Our results confirm that weighted vector averaging can outperform context-sensitive models in most benchmarks, but structural features encoded in RNN models can also be useful in certain classification tasks. We analyzed some of the evaluation datasets to identify the aspects of meaning they measure and the characteristics of the various models that explain their performance variance.  

 \end{abstract}

\section{Introduction}

\blfootnote{
    %
    %
    %
    %
    %
    %
    \hspace{-0.65cm}  
     This work is licensed under a Creative Commons 
     Attribution 4.0 International License.
     License details:
     \url{http://creativecommons.org/licenses/by/4.0/}
}

Distributed semantic models for words encode latent features that reflect semantic aspects and correlations among words. The goal of compositional semantic models is to induce latent semantic representations that encode the meaning of phrases, sentences, and paragraphs of variable lengths. Some neural architectures such as convolutional \cite{kim2014convolutional} and recursive networks \cite{socher2013recursive} handle variable-length input by identifying shift-invariant features suitable for the classification problem at hand, which makes it possible to skip composition and work directly with the entire space of individual word embeddings. While such models can achieve excellent performance in supervised classification tasks such as sentiment analysis, we are interested in generic unsupervised fixed-length representations for variable-length text sequences so as to efficiently preserve essential semantic content for later use in various supervised and unsupervised settings. 

Binary bag-of-words are simple and effective representations that serve as a strong baseline in several classification benchmarks \cite{wang2012baselines}. However, they do not exploit the distributional relationships among different words, which limits their applicability and generalization when training data are scarce. Additive compositional functions, such as word vector sum or average, are more effective in semantic similarity tasks even when compared with tensor-based compositional functions \cite{milajevs2014evaluating} and can outperform more complex and better tuned models based on recurrent neural architectures on out-of-domain data \cite{wieting2015towards}.  Yet, averaging also has several drawbacks: unlike binary representations, the individual word identities are lost, and some words that do not carry semantic significance may end up being more prominently represented than essential words. Furthermore, additive compositional models disregard sentence structure and word order, which can lead to loss of semantic nuance. To alleviate the first issue, the weights of various words can be adjusted using word frequency statistics \cite{riedel2017simple} or by inducing context-sensitive weights using recurrent neural networks \cite{wieting2017revisiting}, both of which have been shown to outperform vector averaging. Context-sensitive feed-forward neural models like the paragraph vector \cite{le2014distributed} potentially incorporate word order, yet the training objective may not be sufficient to model deeper structure. Sequence encoder-decoder models, on the other hand, can be trained with various sentence-level objectives, such as neural machine translation (NMT) \cite{sutskever2014sequence}, predicting surrounding sentences (i.e. skip-thought) \cite{kiros2015skip}, or reconstruction of the input using denoising auto-encoders \cite{hill2016learning}. These sequential models have been evaluated and compared against other models and baselines on several supervised and unsupervised tasks in \newcite{hill2016learning}. The denoising autoencoder model and skip-thought both performed well in supervised tasks, while the NMT model performed worse than the baselines. All three performed poorly in unsupervised settings. 

To bridge some of the gaps in evaluation, we evaluated a subset of models with increasing complexity, from binary bag-of-words to RNNs, on various supervised and unupervised settings. Our objective is to evaluate compositional models against strong baselines and identify the elements that lead to performance gains. We evaluated binary vs. distributed features, weighted vs. unweighted averaging, three different word embedding models,  and four context-sensitive models that optimize different objectives: the paragraph vector, the gated recurrent averaging network \cite{wieting2017revisiting}, skip-thought, and an LSTM encoder trained on labeled natural language inference data (inferSent) \cite{conneau2017supervised}. 
We also analyzed the intrinsic structures of the various models by visual inspection and k-means clustering to gain insights into structural differences that may explain the variance in performance.


\section{Background: Unsupervised Compositional Models}
\subsection{Baselines}
The simplest way of representing a sentence is a  binary bag-of-words representation, where each word is a feature in the vector space. This results in large and sparse representations that only account for the existence of individual words within a sentence, yet they have been shown to be effective in various supervised classification tasks, especially in combination with  \emph{n}-grams and Naive Bayes (NB) features \cite{wang2012baselines}. Let $\vec{x}_i$ be the binary representation of sentence $i$, and $y_i \in \{0,1\}$ its label. The log-count ratio $\vec{r}$ is calculated as 

\begin{equation}
\vec{r}=\log \frac{\vec{p}/{\norm{\vec{p}}}}{\vec{q}/{\norm{\vec{q}}}}
\end{equation}

Where $\vec{p}=1+\sum_{i:y_i =1} \vec{x}_i$ and $\vec{q}=1+\sum_{i:y_i =0} \vec{x}_i$ are the smoothed count vectors for each class (i.e. the number of samples in the class that include each feature). The feature vectors are then modified using the element-wise product $\vec{x}_i \circ \vec{r}$. NB features identify the most discriminative words for each task, so using them results in task-specific rather than general representations. However, given the relative efficiency of this model, we include it as a baseline for comparison. 

\subsection{Word Embeddings and Composition Functions}

Representations of variable-length sentences and paragraphs can be constructed by averaging the embeddings of all words within a sentence. However, simple averaging may not be the best approach since not all words within a sentence are semantically relevant. The following methods can be used to adjust the weights of words according to their frequency, assuming that frequent words have lower semantic content:

\paragraph{\texttt{tf-idf}-weighted Average}

The \textit{term frequency-inverse document frequency} statistic measures the importance of a word to a document.  We treat each sentence as a document and calculate the \texttt{idf} weight for term $t$ as follows: 

\begin{equation}
\textit{idf}_t = \texttt{log} \frac{N}{1+n_t}
\end{equation}

\noindent where N is the total number of sentences and $n_t$  the number of sentences in which the term appears. Terms that appear in more documents have lower \texttt{idf} weights. 

\paragraph{\texttt{sif}-weighted Average}

The \textit{smooth inverse frequency} \cite{riedel2017simple} is an alternative measure for discounting the weights of frequent words as follows:

\begin{equation}
\textit{sif}_t = \frac{a}{a+p(t)}
\end{equation}

\noindent where $a$ is a smoothing parameter and $p(t)$ is the relative frequency of the term in the training corpus. In addition, as proposed in \cite{riedel2017simple}, we subtract the projection of the vectors on the first principal component which corresponds to syntactic features associated with common words. 

\subsubsection{Word Embeddings}
\paragraph{Random word projections:} we generated a random vector drawn from the standard normal distribution for each word in the vocabulary. The vector sum of random word vectors is a low-dimensional projection of binary bag-of-words vectors. 
.  
\paragraph{Continuous Bag of Words:} \texttt{CBOW} is an efficient log-linear model for learning word embeddings using a feed-forward neural network classifier that predicts a word given the surrounding words within a fixed context window \cite{mikolov2013efficient}. In \newcite{schnabel2015evaluation} word embedding evaluation, \texttt{CBOW} outperformed other word embeddings in word relatedness and analogy tasks. 

\paragraph{Global Vectors:} \texttt{GloVe} is a global log-bilinear regression model \cite{pennington2014glove} that produces word embeddings using weighted matrix factorization of word co-occurrence probabilities.

\paragraph{Subword Information Skip-gram} \texttt{si-skip} learns representations for  \emph{n}-grams of various lengths, and words are represented as sums of \emph{n}-gram representations \cite{bojanowski2016enriching}. The learning architecture is based on the continuous skip-gram model \cite{mikolov2013distributed}, which is trained by maximizing the conditional probability of context words within a fixed window with negative sampling. The model exploits the morphological variations within a language to learn more reliable representations, particularly for rare morphological variants. 

\subsection{Neural Compositional Models}
 
Several models have been proposed to overcome some of the weaknesses of bag-of-words and additive representations, such as lack of structure. We evaluated the following context-sensitive models:

\paragraph{The Paragraph Vector:} \texttt{doc2vec} distributed memory model \cite{le2014distributed} constructs representations for sentences and paragraphs using a neural feedforward network that maximizes the conditional probability of words within a paragraph given a context window and the paragraph embedding, which is shared for all contexts generated from the same paragraph. After learning word and paragraph embeddings for the training corpus, the model learns representations for new paragraphs by fixing the model parameters and updating the paragraph embeddings using backpropagation. This additional training at inference time considerably increases the time complexity of the model compared to all others in this study.

\paragraph{Gated Recurrent Averaging Network:} \texttt{GRAN} has been recently introduced to combine the benefits of LSTM networks and averaging, where the weights are computed along with the word and sentence representations \cite{wieting2017revisiting}. The model is trained using aligned sentences that are assumed to be paraphrases to maximize the similarity of their representations against negative examples.  The intuition is to make the averaging operation context-sensitive, resulting in a more powerful construction than simple averaging where all words are equally important. The model was shown to outperform averaging and LSTM models in semantic relatedness tasks. 

\paragraph{Skip-Thought:} The \texttt{skip-th} model is a sequence encoder-decoder trained by projecting sentences into fixed-length vectors, which in turn are used as input to a decoder that is trained to reconstruct surrounding sentences \cite{kiros2015skip}, where the encoder and decoder are RNNs with GRU activations \cite{chung2014empirical}. The model is trained with contiguous sentences extracted from a collection of novels. After training, the model's vocabulary is expanded by learning a linear mapping from pre-trained \texttt{CBOW} word embeddings to the vector space of the \texttt{skip-th} word embeddings. 

\paragraph{Natural Language Inference Encoder:} In \texttt{inferSent} \cite{conneau2017supervised}, a bidirectional LSTM encoder with max-pooling is trained jointly with an inference classifier trained on the Stanford Natural Language Inference (SNLI) dataset \cite{bowmanlarge}, which is a large manually-annotated dataset of English sentence pairs and their inference labels: \{entailment, contradition, neutral\}. 


\section{Evaluation Datasets}

To evaluate the text representations, we used them as features in extrinsic supervised and unsupervised tasks that reflect various semantic aspects, which can be grouped in three categories: pairwise-similarity, sentiment analysis, and categorization. A summary of the dataset statistics is in Table \ref{tab:data}.\footnote{Evaluation scripts and data can be downloaded from: https://github.com/h-aldarmaki/sentence\_eval}
 
\subsection{Pairwise Similarity}

\paragraph{Semantic Textual Similarity:} the STS benchmark dataset \cite{cer2017semeval} includes a collection of English sentence pairs and human-annotated similarity scores that range from 0 (unrelated sentences) to 5 (paraphrases). The dataset includes training, development, and test sets. This task can be performed without supervision by calculating the cosine similarity between two sentence vectors. We also evaluated the models in a  supervised settings using linear regression, where the input vector is a concatenation of the element-wise produce $u.v$ and absolute difference $\vert u-v \vert$ of each pair $\langle u,v \rangle$.

\paragraph{Sentences Involving Compositional Knowledge:} SICK dataset is a benchmark for evaluating compositional models \cite{marelli2014sick}. We evaluated the models on the relatedness subtask, which is constructed in a similar manner as STS benchmark. 

\paragraph{Paraphrase Detection:} This is a binary classification task that involves the identification of paraphrases in similar sentence pairs using the Microsoft Research Paraphrase Corpus, MSRP \cite{dolan2004unsupervised}. We evaluated the models in two ways: calculating the cosine similarity between the sentence pairs and classifying them as paraphrases if the similarity is larger than a threshold tuned from the training set. The second approach is to learn a logistic regression classifier using a concatenation of  $u.v$ and $\vert u-v \vert$. 

\subsection{Sentiment Analysis and Text Categorization}

\paragraph{Sentiment Analysis:} We used the following binary classification tasks: \textbf{CR} customer product reviews \cite{hu2004mining}, \textbf{MPQA} opinion polarity subtask \cite{wiebe2005annotating}, \textbf{RT-s} short movie reviews \cite{pang2005seeing}, \textbf{Subj} subjectivity/objectivity classification task \cite{pang2004sentimental}, and \textbf{IMDB} full-length movie review dataset \cite{maas2011learning}.

\paragraph{Newsgroups:} Following the setup in \cite{wang2012baselines} we used the 20-Newsgroup dataset \footnote{http://qwone.com/$\sim$jason/20Newsgroups/} to extract several binary topic categorization tasks. We processed the datasets to remove headers, forwarded text, and signatures, which results in smaller sentences and paragraphs. We used the following newsgroups for binary classification: \textbf{religion} (atheism vs. religion), \textbf{sports} (baseball vs. hockey), \textbf{computer} (windows vs. graphics), and \textbf{politics} (middle east vs. guns). We also trained multi-class classifiers on the 8 newsgroups.

\paragraph{Question Classification:} We used the TREC 10 coarse question categorization task \footnote{http://cogcomp.org/Data/QA/QC/} which categorizes questions into 6 classes: human (HUM), entity (ENTY), location (LOC), number (NUM), description (DESC),  and abbreviation (ABBR).

\begin{table}
\resizebox{\columnwidth}{!}{
\begin{tabular}{|c|c|c|c|c|c|c|c|c|c|c|c|c|c|}
\hline
\multirow{2}{*}{Dataset} & \multicolumn{3}{c|}{Pair-wise Similarity} & \multicolumn{5}{c|}{Sentiment Analysis} &\multicolumn{4}{c|}{Newsgroup } & \multirow{2}{*}{TREC}\\
\cline{2-13}
& STS & SICK & MSRP & CR & MPQA & RT-s & Subj & IMDB & REL & SPO & COM & POL & \\
\hline
Train & 5,749 & 4,934 & 4,076 & 3,775 & 10,606 & 10,662 & 10,000 & 25k& 1,078 & 1,604 & 1,694 & 1,310 & 5,452\\
\hline
Test & 1,379 & 4,906 & 1,725 & -- & -- & -- & --& 25k & --& -- & -- & -- & 500\\
\hline
pos ratio & -- & -- & 0.66 & 0.64 & 0.31 & 0.50 & 0.50 & 0.50 & 0.58 & 0.51 & 0.51 & 0.52 & --\\
\hline
$l$ & 12 & 10 &  23 & 21 & 3 & 21 & 24 & 262 & 82 & 82 & 82 & 92 & 10 \\
\hline
\end{tabular}
}
\caption{Dataset statistics. \textbf{Train}: number of samples in the training set. \textbf{Test}: number of samples in the test set, if applicable (CV is applied otherwise). \textbf{pos ratio}: ratio of positive samples in the test set (or total for datasets with no splits). \textbf{$l$}: average length of all samples. }\label{tab:data}
\end{table}


\section{Experimental Setup}

\subsection{Training Data}

We trained the unsupervised word embedding models \texttt{CBOW},  \texttt{GloVe}, and \texttt{si-skip} on a set of $\sim$ 7 million sentences extracted from the English Wikipedia and Amazon movie and product reviews \cite{he2016ups}. We also trained the Paragraph Vector (\texttt{doc2vec}) model on this dataset, and initialized the word embeddings using the \texttt{si-skip} pre-trained word embeddings above. While better results overall could be obtained using pre-trained word embeddings trained with much larger text corpora, we used this medium-size corpus to evaluate the various models consistently and reduce model variability due to data and vocabulary coverage. 

We used the publicly available pre-trained \texttt{GRAN}\footnote{https://github.com/jwieting/acl2017}  and \texttt{skip-th}\footnote{https://github.com/ryankiros/skip-ths}  models, which require training with special types of datasets: paraphrase collections, and contiguous text from books, respectively. To ensure a fair evaluation, we only compared these models against binary bag-of-words and equivalent word embeddings. The word embeddings within the \texttt{GRAN} model were initialized with \texttt{PARAGRAM-SL999} word vectors \cite{wieting2015paraphrase}, so we used them as an evaluation baseline for \texttt{GRAN}. We compared \texttt{skip-th} against the \texttt{CBOW} embeddings that were used to expand the vocabulary, which account for most words in the final model's vocabulary. We used the pre-trained  \texttt{inferSent} model \footnote{https://github.com/facebookresearch/InferSent} which uses pre-trained GloVe word embeddings \footnote{https://nlp.stanford.edu/projects/glove}. We also experimented with the post-trained word embeddings for each model with similar results, so we omitted them for brevity.  

\subsection{Training Settings}
We trained the unsupervised word embedding models using the optimal parameters recommended for each model. The hyper-parameters in \texttt{doc2vec} were set according to the recommendations in \cite{lau2016empirical}. For the supervised sentiment classification and text categorization tasks, we trained and tuned linear SVM models using grid search for datasets that include train/dev/test splits, and nested cross-validation otherwise. We also experimented with kernel SVMs but didn't observe notable differences in the results. 

\section{Evaluation Results}

\subsection{Pairwise Similarity Evaluation} 

\hspace{-10pt}
\begin{figure}[H]
\vspace{-10pt}
\begin{minipage}{0.195\textwidth}
 \centering
 \small $0.5$\\
    \includegraphics[width=\textwidth]{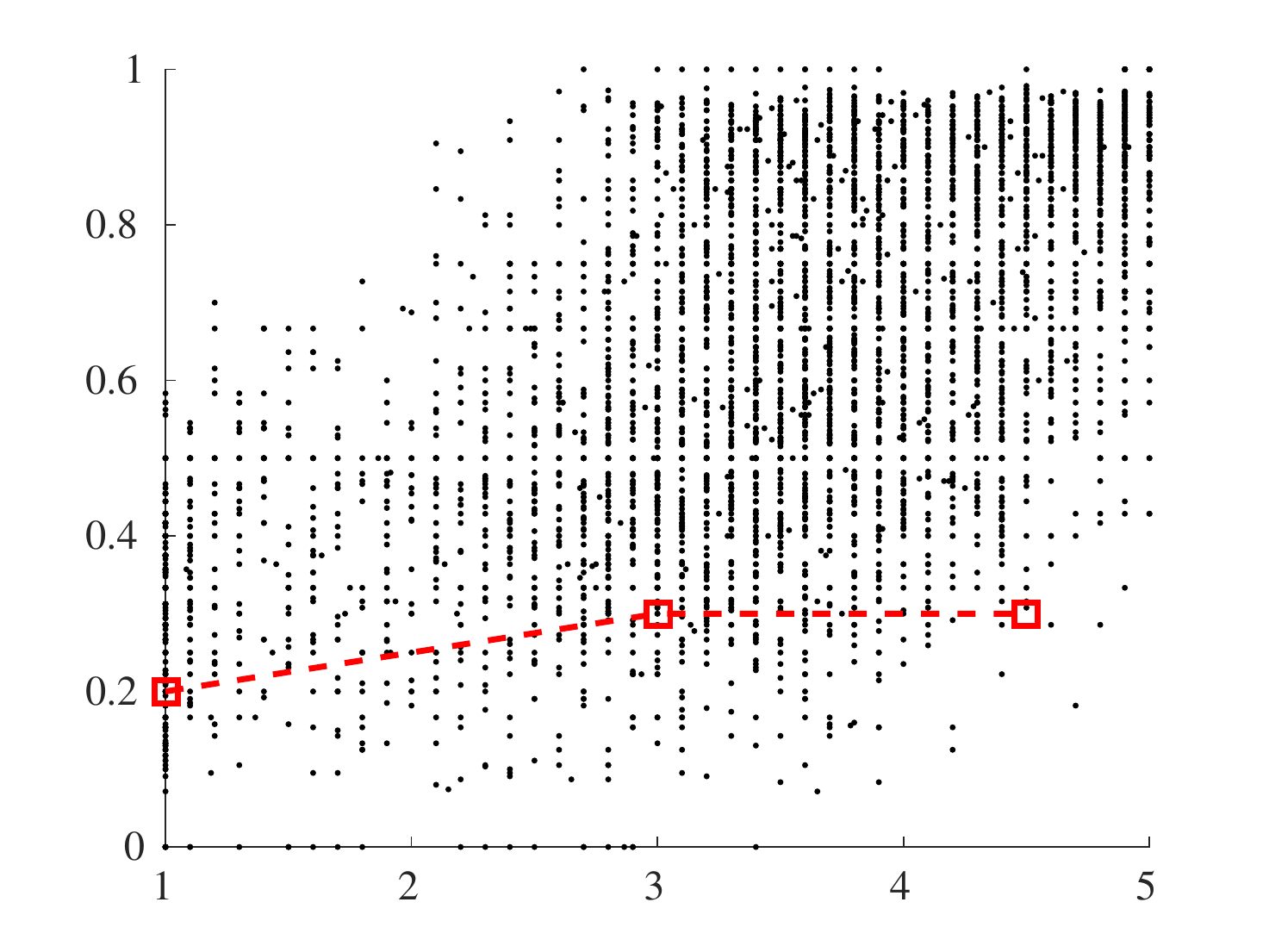}
\end{minipage}
\begin{minipage}{0.195\textwidth}
 \centering
   \small  $0.65$\\
    \includegraphics[width=\textwidth]{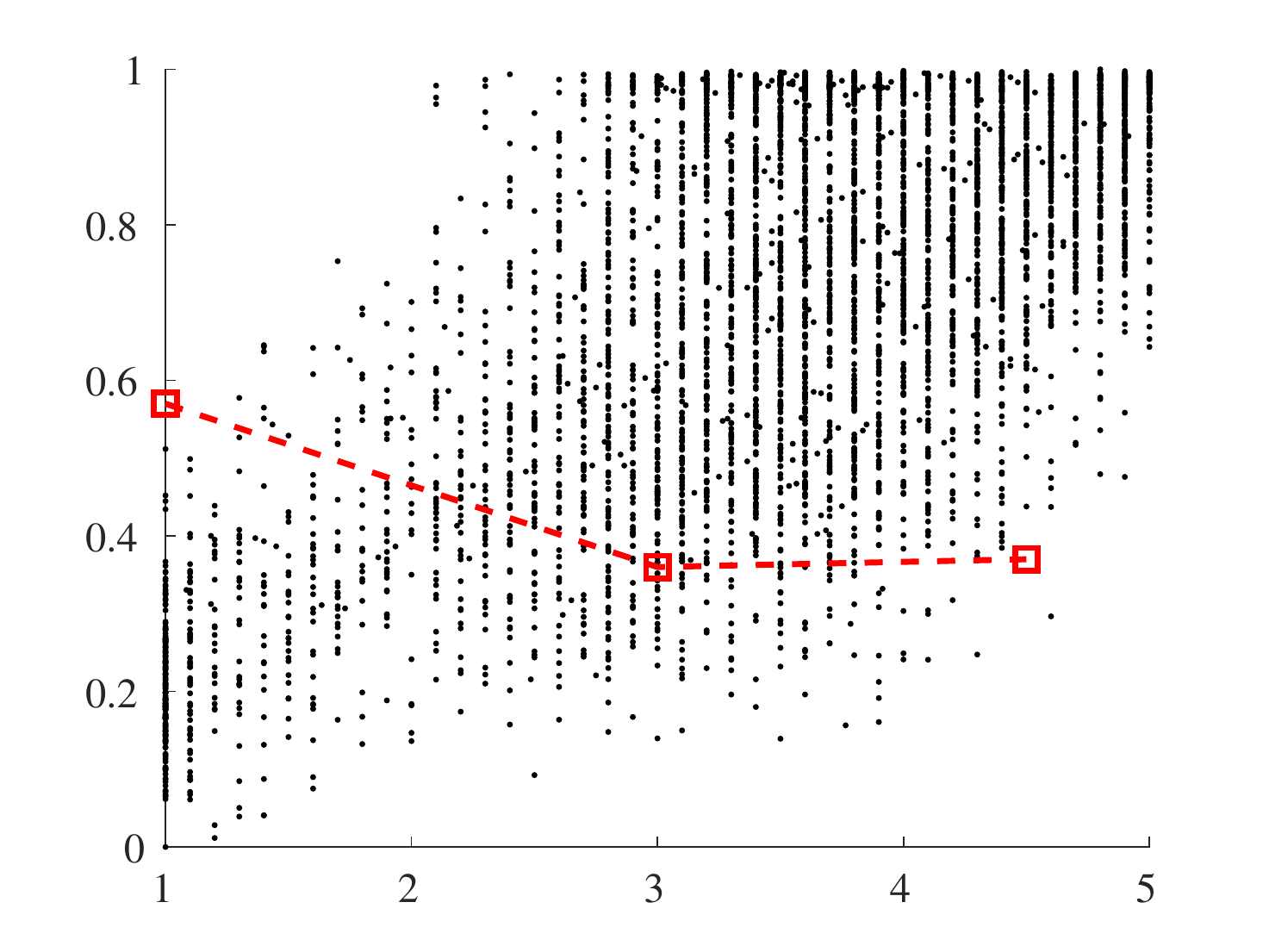}
\end{minipage}
\begin{minipage}{0.195\textwidth}
 \centering
  \small $0.5$\\
    \includegraphics[width=\textwidth]{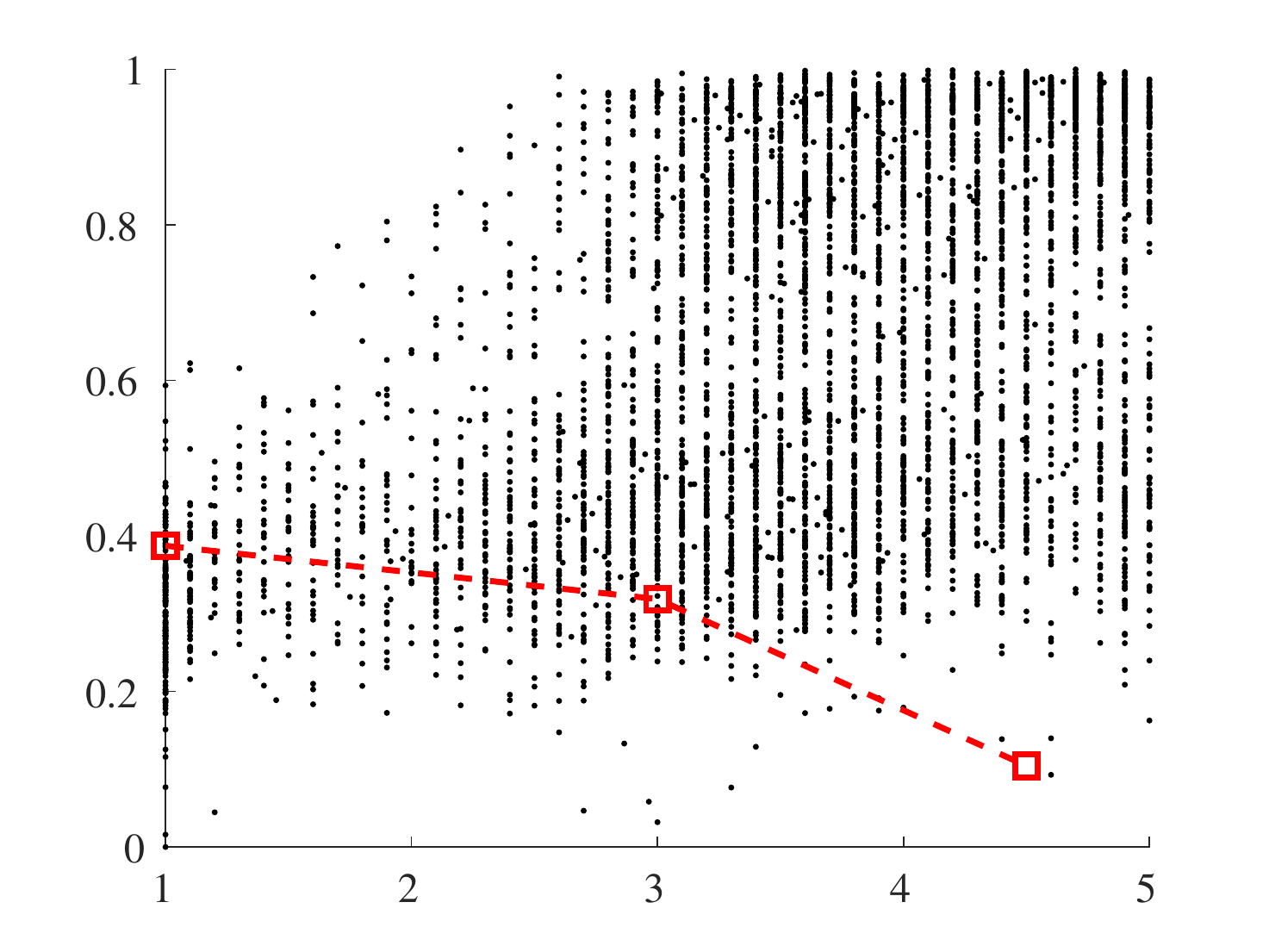}
\end{minipage}
    \begin{minipage}{.195\textwidth}
    \centering
      \small $0.73$\\
    \includegraphics[width=\textwidth]{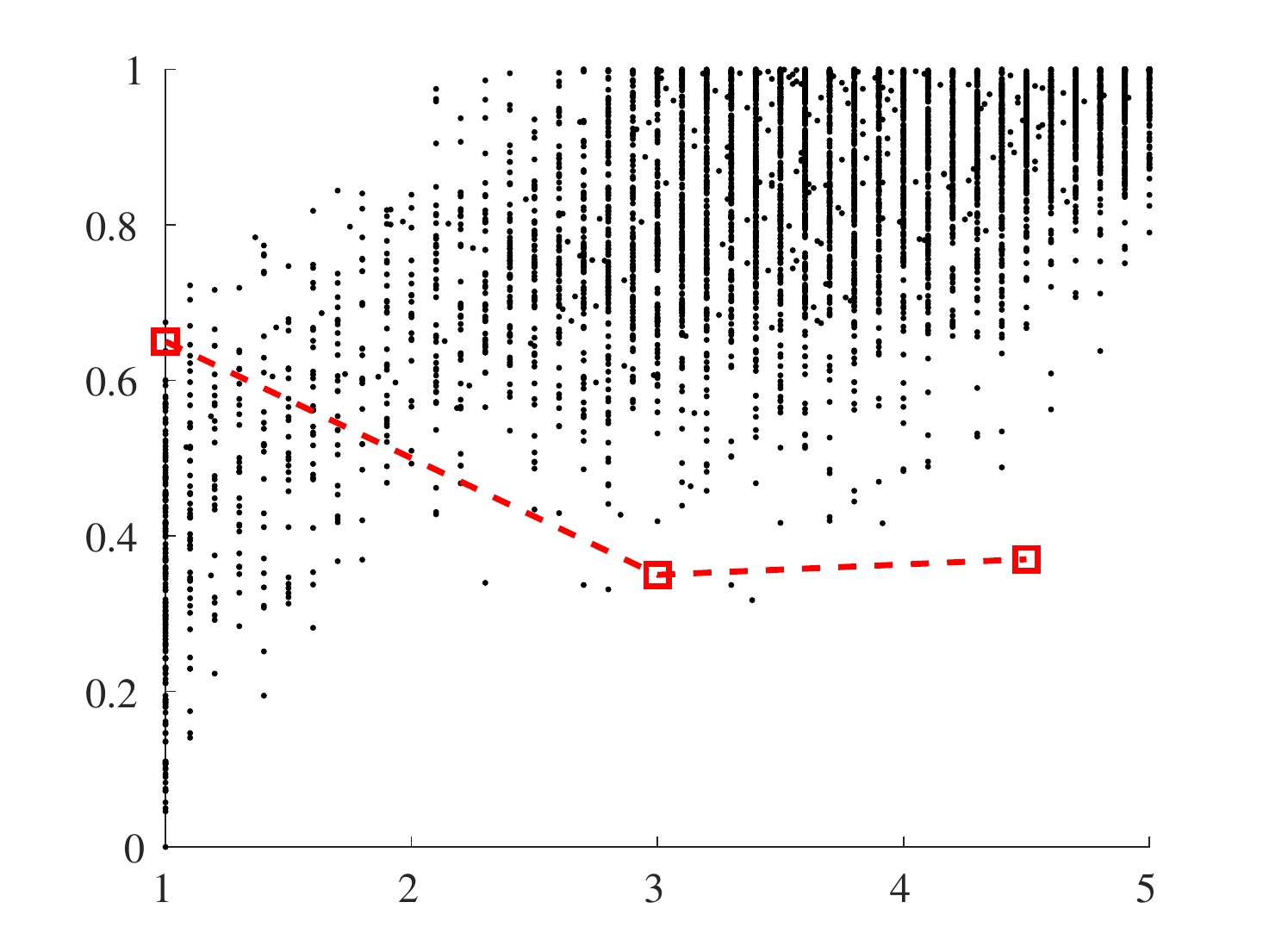}
    \end{minipage}
        \begin{minipage}{.195\textwidth}
    \centering
    \small  $0.74$\\
    \includegraphics[width=\textwidth]{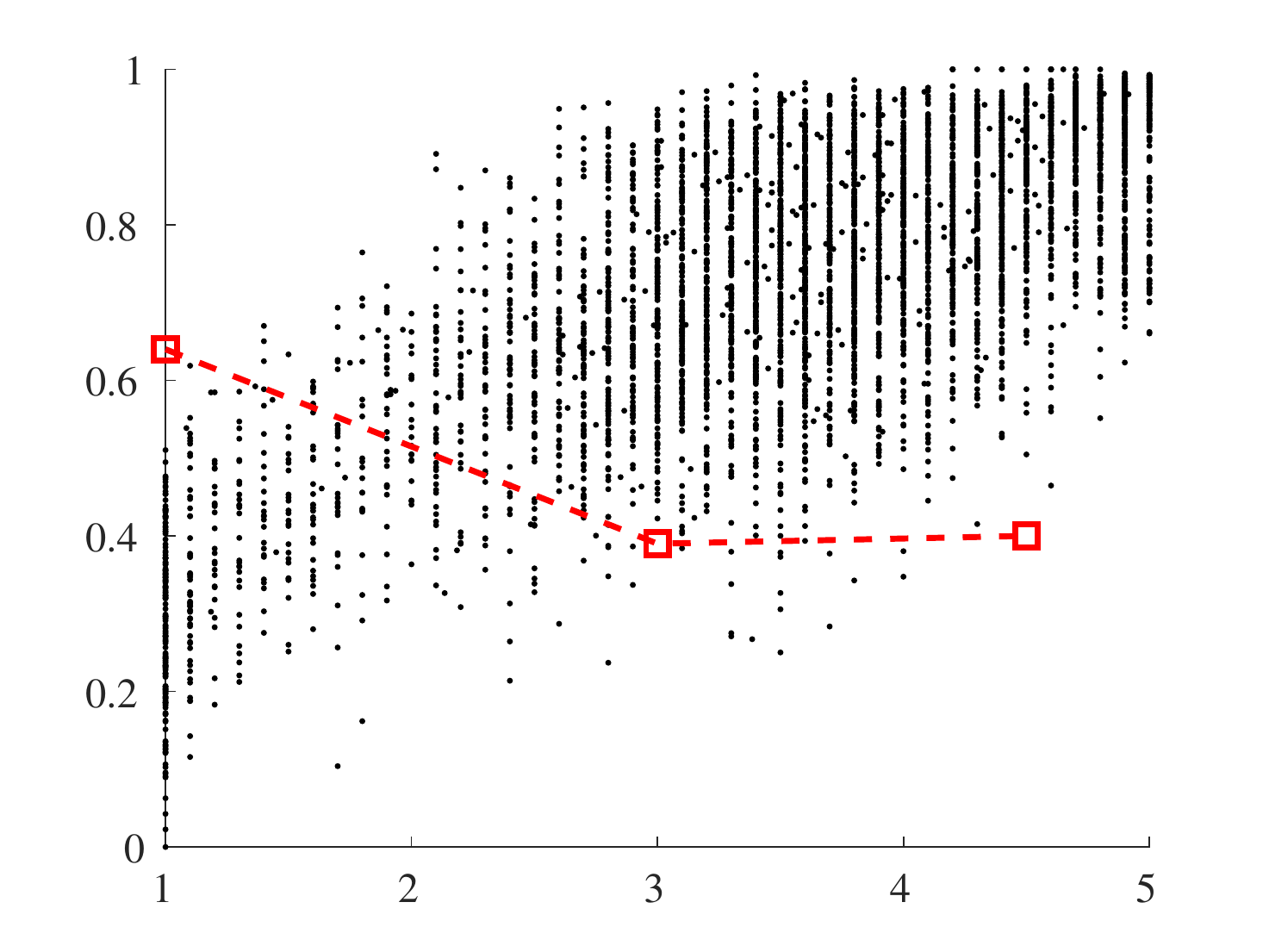}
    \end{minipage}\\
    
\begin{minipage}{0.195\textwidth}
 \centering
   \small $0.6$\\
    \includegraphics[width=\textwidth]{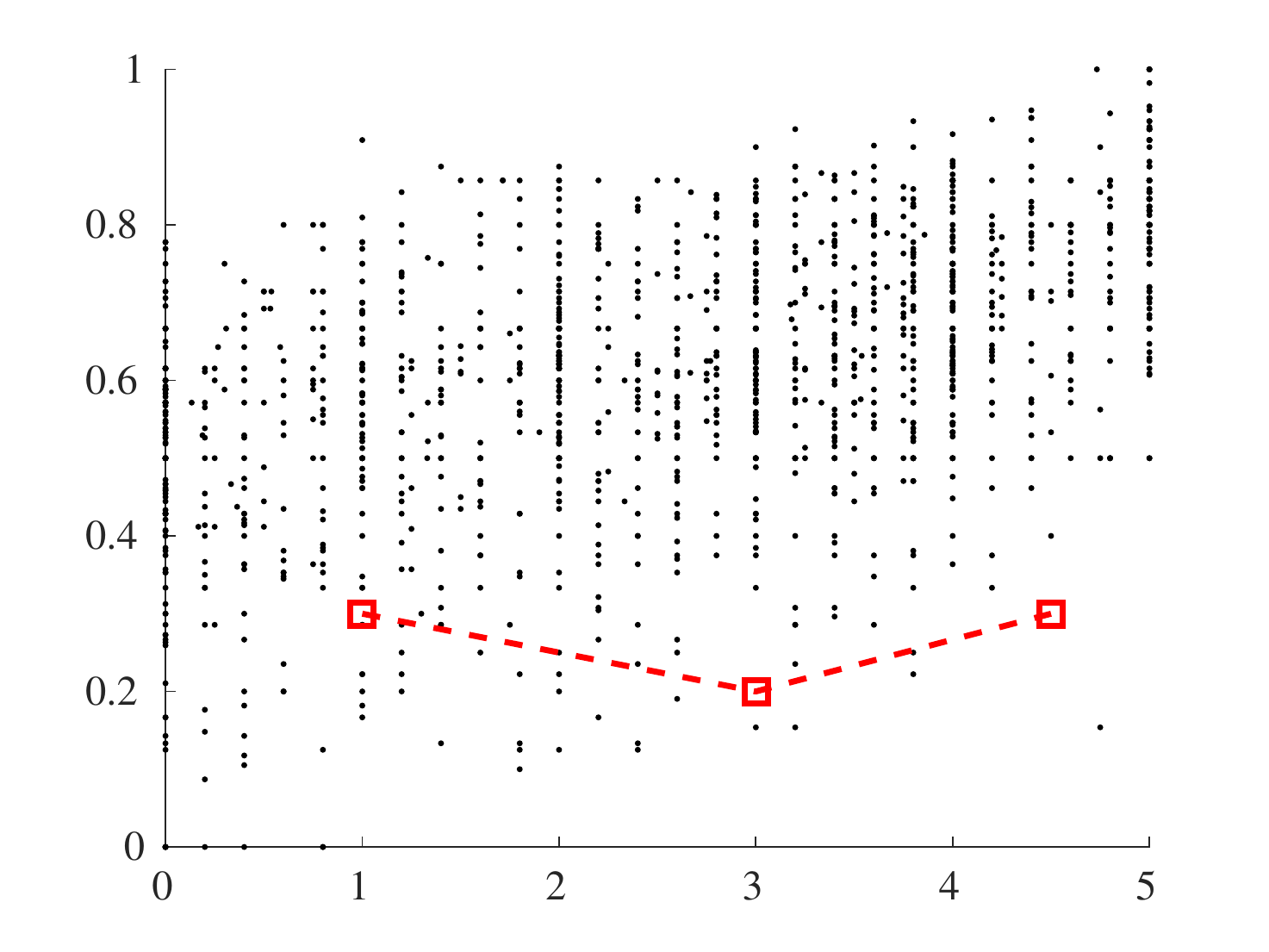}
       \captionof{subfigure}{\small word overlap}
\end{minipage}
\begin{minipage}{0.195\textwidth}
 \centering
   \small $0.63$\\
    \includegraphics[width=\textwidth]{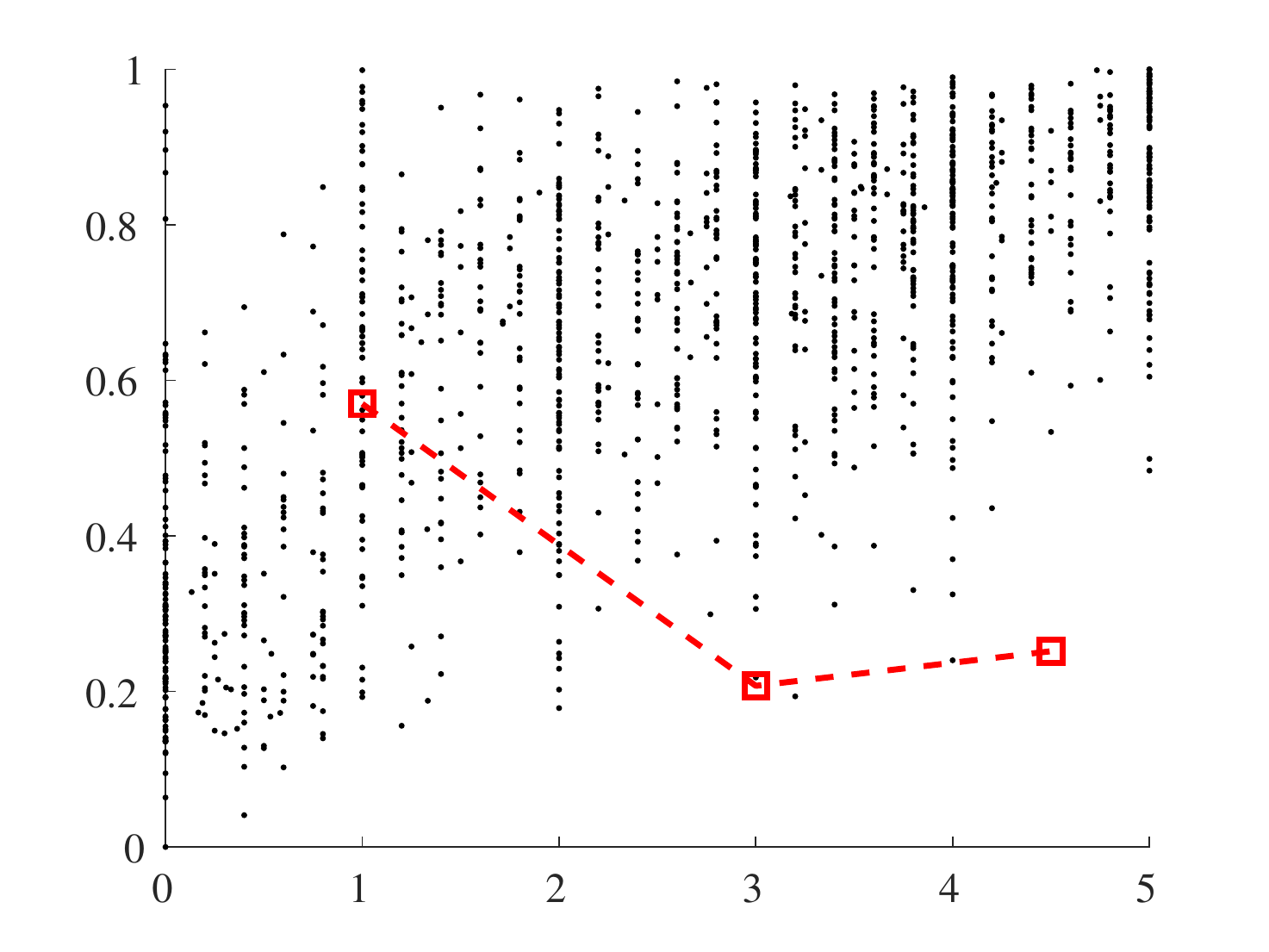}
       \captionof{subfigure}{\small doc2vec}
\end{minipage}
\begin{minipage}{0.195\textwidth}
 \centering
   \small $0.2$\\
    \includegraphics[width=\textwidth]{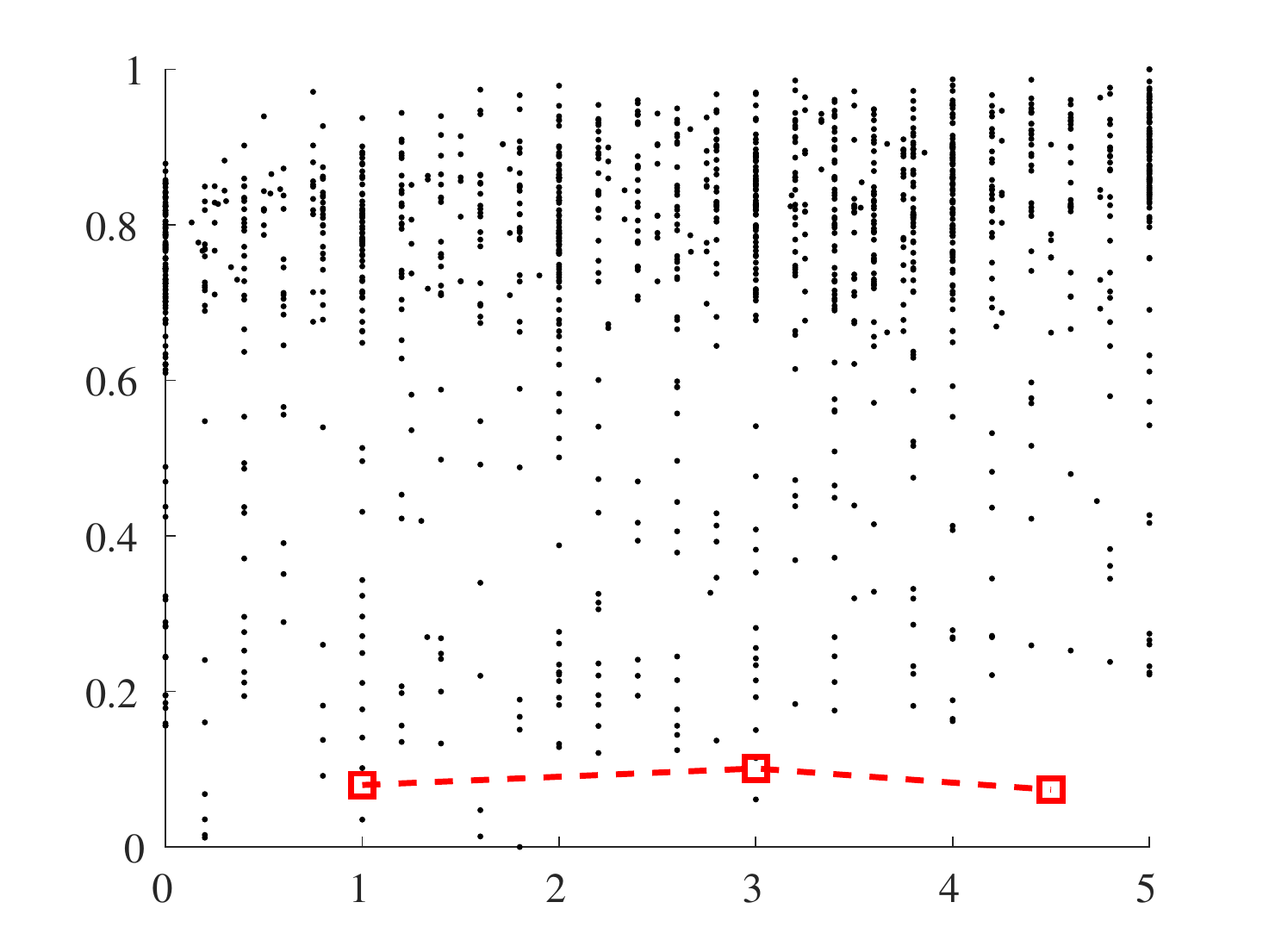}
      \captionof{subfigure}{\small skip-th}
\end{minipage}
    \begin{minipage}{.195\textwidth}
    \centering
     \small $0.68$\\
    \includegraphics[width=\textwidth]{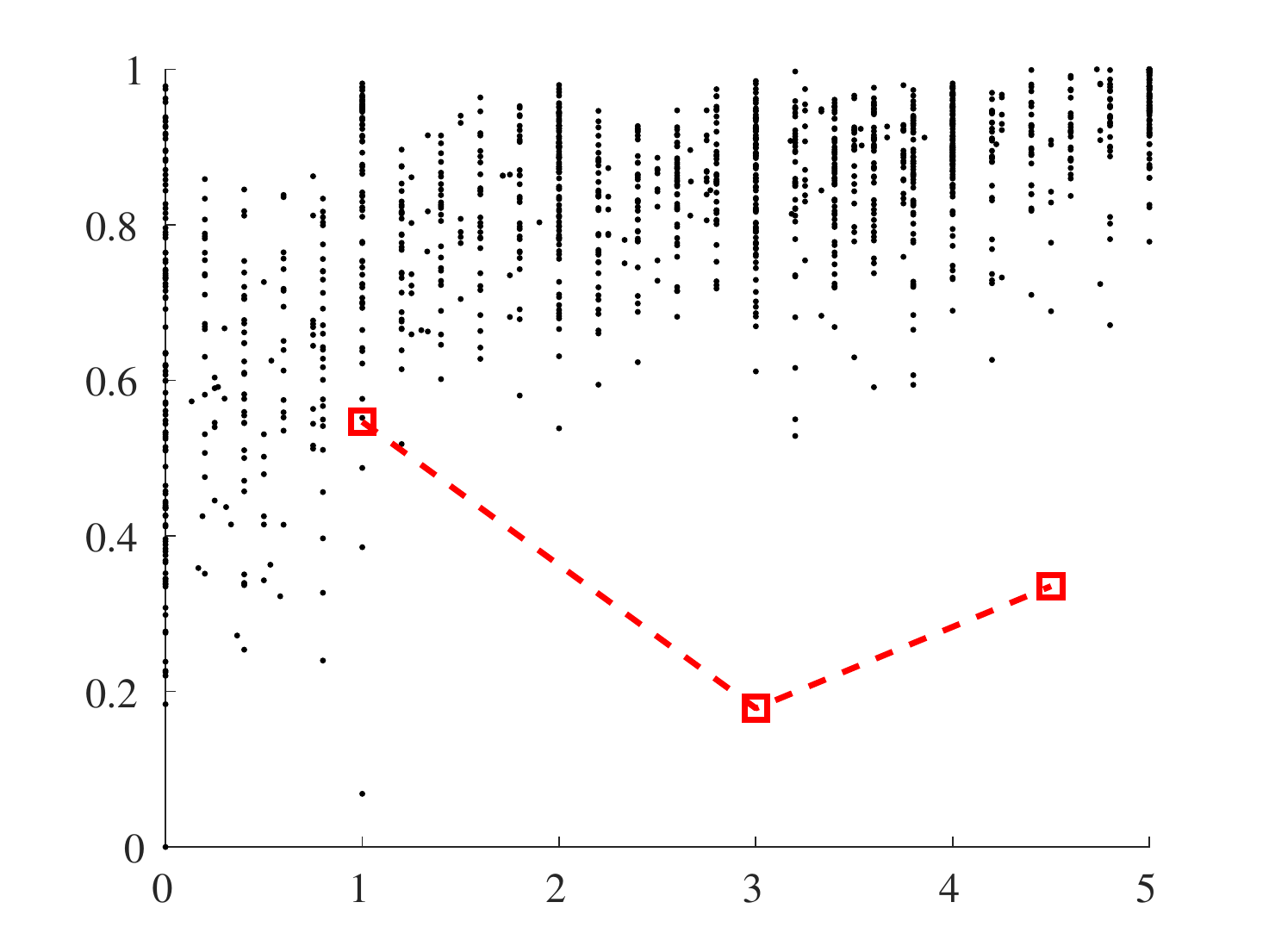}
      \captionof{subfigure}{\small sif$\dagger$}
    \end{minipage}
        \begin{minipage}{.195\textwidth}
    \centering
     \small $0.69$\\
    \includegraphics[width=\textwidth]{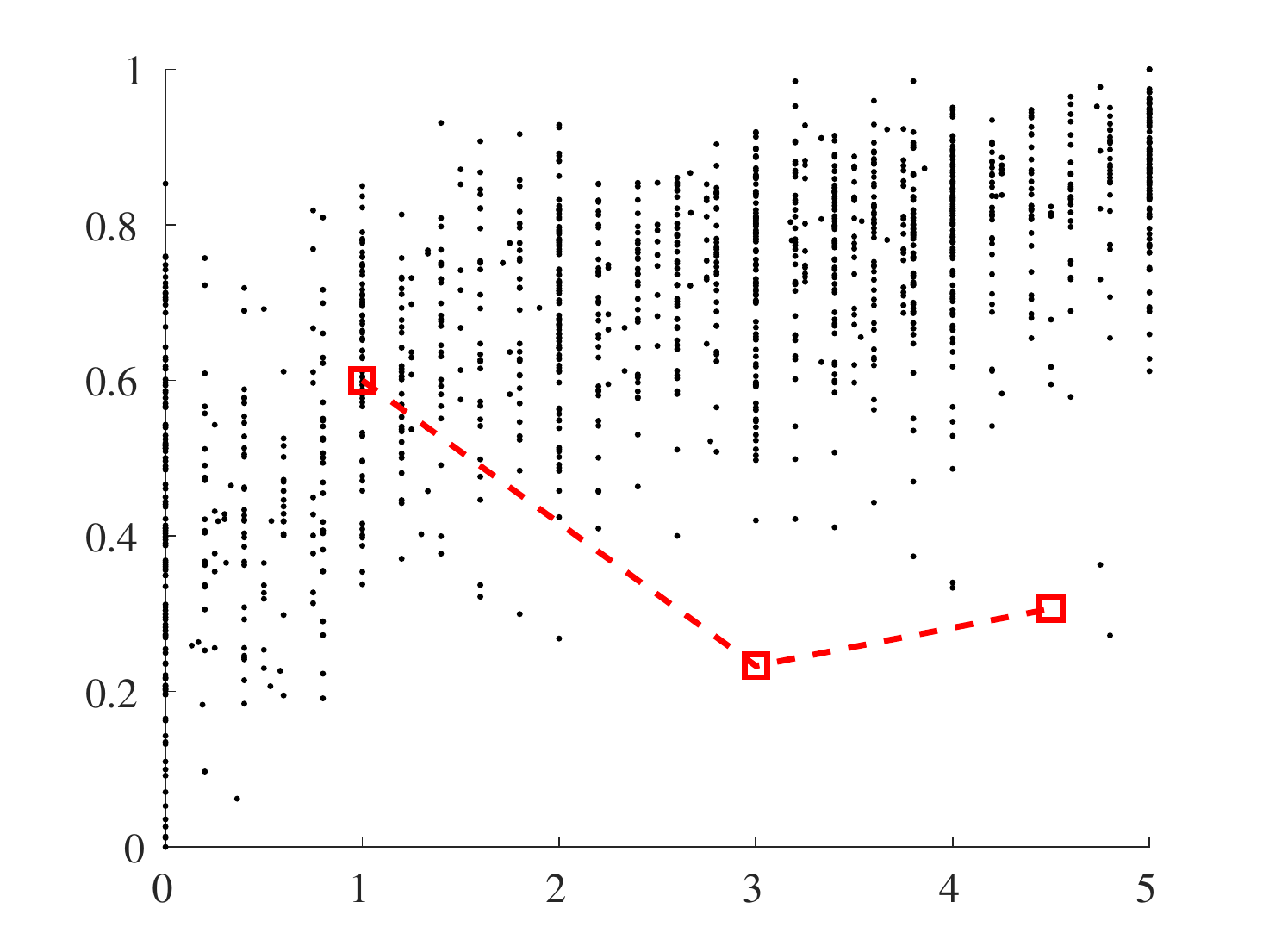}
      \captionof{subfigure}{\small inferSent}
    \end{minipage}
    \caption{Scatter plots of normalized gold scores in the x axis vs. (a) word overlap (\%) and  (b - e) cosine similarity using various models. Top: SICK. Bottom: STS Benchmark. Pearson $\rho$ plotted in red for  $score \le 2$, $score \in (2,4)$, and $score \ge 4$. Overall Pearson $\rho$ shown at the top. $\dagger$ \small sif-weighted average of pre-trained Glove vectors used in inferSent.}\label{fig:corr}
\end{figure}

\begin{table}[th]
\small
\centering
\resizebox{0.8\columnwidth}{!}{
  \begin{tabular}{|c|c|c|c|c|c?c|c|}
    \cline{3-8}
    \multicolumn{2}{l|}{} & 
    \multicolumn{2}{c|}{STS Benchmark $\rho$ } & \multicolumn{2}{c?}{SICK  $\rho$} & 
    \multicolumn{2}{c|}{MSRP accuracy/F1} \\
    \cline{3-8}
    \multicolumn{2}{l|}{} &
    \multicolumn{1}{c|}{cosine} & \multicolumn{1}{c|}{linear reg.} & \multicolumn{1}{c|}{cosine} & \multicolumn{1}{c?}{linear reg.} & \multicolumn{1}{c|}{cosine} &  \multicolumn{1}{c|}{logistic reg.}  \\
    \hline
    \multicolumn{2}{|c|}{Binary BOW}  & 0.536 &0.606 & 0.611 & \BW 0.761 &66.9/0.765 & \G 72.2/0.811 \\
     \hline
     \specialrule{.2em}{.1em}{.1em} 
           \multicolumn{2}{|c|}{Paragraph Vector (doc2vec)}  & 0.628 & 0.673 & 0.654 & 0.655 & 68.6/0.797 &70.4/0.803 \\
     \hline
         \multirow{3}{*}{Random}  & avg & 0.558 & 0.616  & 0.602 & 0.669 & \G 70.6/0.780 & 70.8/0.797\\
    \cline{2-8}
    &idf& \GG 0.668 & 0.665  & 0.617 & 0.659 & \G 70.0/0.790 &69.1/0.791\\
    \cline{2-8}
    &sif& \GG 0.666 & 0.665 & 0.628 & 0.655 & \G 70.1/0.786 &69.9/0.699 \\
    \hline
      \multirow{3}{*}{CBOW}  & avg & 0.630 & 0.672  & 0.679 & \G 0.728 & \G 71.9/0.815 &\G 72.0/0.807\\
    \cline{2-8}
    &idf& \GG 0.697 &  \G 0.695  & 0.678 & \G 0.712 & \G 71.8/0.815 &\G 72.3/0.809\\
    \cline{2-8}
    &sif& \GG 0.683 & 0.686  & 0.690 &\G 0.715& \G 72.2/0.814 &\G 71.4/0.804\\
    \hline
    \multirow{3}{*}{GloVe}  & avg & 0.336 & 0.574 & 0.602 & 0.694 &68.9/0.807 &\G 71.0/0.810 \\
    \cline{2-8}
    &idf& 0.540 & 0.656 & 0.624 & 0.685 & \G 71.3/0.818 &\G 73.4/0.820 \\
    \cline{2-8}
    &sif& \GG 0.685 & 0.665  & \GG 0.701 & 0.695 & \G 71.8/0.809 &\G 72.1/0.811\\
    \hline
    \multirow{3}{*}{si-skip}  & avg & 0.608 & \G 0.690 & 0.684 &\G 0.730& \G 72.1/0.817 &\G 71.9/0.895 \\
    \cline{2-8}
    &idf& \GG 0.683 & \GG 0.714 & \GG 0.702 & \G 0.715 &69.3/0.809 &70.3/0.815 \\
    \cline{2-8}
    &sif& \GG 0.694 & \GG 0.721 & \GG 0.716 & \G 0.721& \G 70.6/0.804 &70.6/0.802 \\
    \hline
    \specialrule{.2em}{.1em}{.1em} 
         \multicolumn{2}{|c|}{GRAN}  &\GG 0.747 & \GG 0.747 & \G 0.715 & \GG 0.756 &71.3/0.817 &72.3/0.812 \\
    \hline
    \multirow{3}{3cm}{\centering Pre-trained PARAGRAM-SL999$\dagger$ }  & avg &0.564 &0.690 &0.694 &  0.746&71.8/0.818 &73.2/0.816 \\
    \cline{2-8}
    &idf& 0.711 & \GG 0.733 & \GG 0.723 & \GG 0.756 &72.1/0.817 &73.3/0.817 \\
    \cline{2-8}
    &sif&  0.716	& \GG 0.722	& \GG 0.733	& \GG 0.765	&  73.4/0.822 &	72.0/0.809\\
    \hline
    \specialrule{.2em}{.1em}{.1em} 
              \multicolumn{2}{|c|}{skip-th}  & 0.213 & \GG 0.729  & 0.498 & \BW 0.811 &62.3/0.761 & 73.0/0.812\\
              \hline
     \multirow{3}{1.8cm}{\centering Pre-trained CBOW$\dagger$ }   & avg & 0.631 & 0.695 &\BW 0.727 & 0.758 &\G 70.3/0.813 &  73.2/0.813 \\
    \cline{2-8}
    &idf& \GG 0.674 &\GG 0.708 & \GG 0.710 & 0.731& \G 69.6/0.809 &71.3/0.807\\
    \cline{2-8}
    &sif& \GG 0.686 &	 \GG 0.707&	\BW 0.727&	0.737	& \G 69.8/0.809&	70.9/0.803\\
    \hline
    \specialrule{.2em}{.1em}{.1em} 
              \multicolumn{2}{|c|}{inferSent}  & \GG 0.692	& \BW 0.773	& \GG 0.744	& \BW 0.865	& 0.697/0.806	& \GG 0.746/0.827\\
              \hline
     \multirow{3}{1.8cm}{\centering Pre-trained GloVe$\dagger$ }   & avg &0.497	&0.655	& 0.687	&0.753	& \G 0.711/0.818	& \G 0.732/0.817 \\
    \cline{2-8}
    &idf& 0.606	& 0.688	& 0.696	& 0.736	& 0.688/0.809	& 0.711/0.804\\
    \cline{2-8}
    &sif& \GG 0.679	& 0.699	& \G 0.729	& 0.749	& \G 0.709/0.816	& 0.708/0.804\\
    \hline
  \end{tabular}
}
\caption{Pearson $\rho$ for STS Benchmark and SICK relatedness, and Accuracy\%/F1 for MSR Paraphrase detection. Results are shaded according to their statistical significance using the Williams test \cite{graham2014testing} with $\alpha = 0.05$. \small $\dagger$ pre-trained vectors used in the model above.} \label{table:sts}  
\end{table}
\if{false}
\begin{table}[h]
\small
\centering
    \scalebox{0.9}{
    \begin{tabular}{p{5.5cm}p{5.8cm}cccc}
       Sentence 1 & Sentence 2 &  Score & Binary & sif & doc2vec\\ 
        \hline
       \textbf{A} person \textbf{is} \underline{\smash{frying}} some \underline{food}. & There \textbf{is} no \textbf{person}  \underline{\smash{peeling}} a \underline{\smash{potato}}. & 0.25 & 0.41 & 0.49& 0.41\\
        \hline
           \underline{A woman} \textbf{is} \underline{\smash{cutting a fish}}. & \underline{The main} \textbf{is} \underline{\smash{slicing potatos}}. & 0.25 & 0.13 & 0.46 & 0.42    \\
         \hline
          \textbf{Two} \underline{\smash{groups of people}} \textbf{are} \underline{\smash{playing}} \bf{football} &  \textbf{Two} \underline{team} \textbf{are} \underline{\smash{competing}} in a \textbf{football} match. & 0.93 & 0.52 & 0.74 & 0.72\\
          \end{tabular}
      }
      \captionof{table}{Examples of sentence pairs in SICK and their relatedness scores vs. cosine similarity scores. All scores were normalized to be in [0,1]. Shared words are shown in bold and related words underlined.} \label{tab:ex1}
\end{table}
\fi
\begin{table}[]
\small
\centering
\hspace{-9.5pt}
    \scalebox{0.85}{
    \setlength{\tabcolsep}{3pt}
    \begin{tabular}{p{5.5cm}p{5.8cm}ccccccc}
       Sentence 1 & Sentence 2 &  Score & Binary & si-sif$\dagger$ & doc2vec & skip-th & infer & gl-sif$\dagger$\\ 
        \hline
       \textbf{A} person \textbf{is} \underline{\smash{frying}} some \underline{food}. & There \textbf{is} no \textbf{person}  \underline{\smash{peeling}} a \underline{\smash{potato}}. & 0.25 & 0.41 & 0.49 & 0.41 & 0.51 & 0.67 & 0.71\\
        \hline
           \underline{A woman} \textbf{is} \underline{\smash{cutting a fish}}. & \underline{The main} \textbf{is} \underline{\smash{slicing potatos}}. & 0.25 & 0.13 & 0.46 & 0.42  & 0.53 & 0.65 & 0.57 \\
         \hline
          \textbf{Two} \underline{\smash{groups of people}} \textbf{are} \underline{\smash{playing}} \bf{football} &  \textbf{Two} \underline{team} \textbf{are} \underline{\smash{competing}} in a \textbf{football} match. & 0.93 & 0.52 & 0.74 & 0.72 & 0.60 & 0.79 & 0.71\\
          \hline
          Different \textbf{teams are playing} \underline{football} on the field & Two \textbf{teams are playing} \underline{soccer} & 0.70 & 0.56 & 0.93 & 0.89 & 0.51 & 0.82 & 0.91\\
          \end{tabular}
      }
      \captionof{table}{Examples of sentence pairs in SICK and their relatedness scores vs. cosine similarity scores. All scores were normalized to be in [0,1]. Shared words are shown in bold and related words underlined. \small $\dagger$ sif-weighted average of \texttt{si-skipgram} and pre-trained \texttt{GloVe}} \label{tab:ex1}
\end{table}

Table \ref{table:sts} shows the performance of the various models in the pair-wise similarity tasks. We highlight the best performance in each block; differences within the same shade are not statistically significant. Among the word embedding models, the subword skipgram \texttt{si-skip} achieved the best overall performance. For all models except \texttt{si-skip}, simple averaging performed poorly in semantic relatedness tasks, especially in the unsupervised setting, while \texttt{sif}-weighting generally outperformed \texttt{idf}-weighting (the improvement is most evident for \texttt{GloVe}). A similar trend is observed with random word vectors, which performed on par with  \texttt{doc2vec}. 

The binary bag-of-words model performed particularly well in the supervised SICK task. The performance of binary and random vectors can be explained by the high correlation between the percentage of overlapping words and the similarity scores as seen in Figure \ref{fig:corr}. We also highlighted the Pearson correlation coefficients for the following subsets of relatedness scores: $A=\{score \le 2\}$, $B=\{score \in (2,4)\}$, and $C=\{score \ge 4\}$. The overall Pearson correlation mostly reflects the performance on the most and least similar pairs, which tend to be the pairs with the highest and lowest word overlap, respectively; within the regions we highlighted, all correlations were relatively low. However, distributed models like \texttt{sif} and \texttt{inferSent} improved the correlation of $A$ for SICK, and both $A$ and $C$ for STS Benchmark. Table \ref{tab:ex1} shows some examples of sentence pairs from $A$ and $C$; while both \texttt{sif} and \texttt{doc2vec} vectors consistently identified similar concepts (namely food related and competition concepts) regardless of surface similarity, binary scores only reflected the lexical similarity, which resulted in inconsistent scores.

\texttt{GRAN} outperformed simple averaging in both the STS and SICK tasks, which confirms the results in \cite{wieting2017revisiting}, but compared with \texttt{idf} and \texttt{sif} averaging, there is no apparent improvement; it only outperformed weighted averaging in the unsupervised STS benchmark. \texttt{skip-th} vectors performed poorly in the unsupervised similarity tasks, but outperformed the pre-trained vectors in the supervised similarity tasks, particularly in SICK. \\

\vspace{-30pt}
\hspace{-15pt}
\begin{figure}[t]
\begin{minipage}{0.67\textwidth}
\small
\centering
    \scalebox{0.8}{
     \begin{tabular}{p{5.5cm}p{5.8cm}c}
      Sentence 1 & Sentence 2 &  Label\\ 
        \hline
         \specialrule{.2em}{.1em}{.1em} 
        \textbf{Bashir felt he was being tried by opinion} not on the \textbf{facts}, Mahendradatta told Reuters.  & \textbf{Bashir} also \textbf{felt he was being tried by opinion} rather than \textbf{facts} of law, he added. & 1\\
        \hline
         \textbf{West Nile Virus}-which \textbf{is spread} through infected \textbf{mosquitoes} -is potentially fatal.  & \textbf{West Nile} is a bird \textbf{virus} that is \textbf{spread} to people by \textbf{mosquitoes}. & 0 \\
                   \hline
          SCO \underline{\smash{says}} the \textbf{pricing} terms for a license \textbf{will} not \underline{be announced} for \underline{week} & Details on \textbf{pricing} \textbf{will} \underline{be announced} within a few \underline{weeks} , McBride \underline{said}  & 1\\       
      \hline
      Russ Britt \textbf{is} the \underline{\smash{Los Angeles}} \textbf{Bureau Chief} for \textbf{CBS.MarketWatch.com.} & Emily Church \textbf{is} \underline{London} \textbf{bureau chief} of \textbf{CBS.MarketWatch.com.} & 0\\
        \end{tabular}
        }
      \captionof{table}{Examples of sentence pairs in MSRP and their labels. Shared words are shown in bold and related words underlined.} \label{tab:ex2}
\end{minipage}\hfill
    \begin{minipage}{.3\textwidth}
    \centering
    \includegraphics[width=\textwidth]{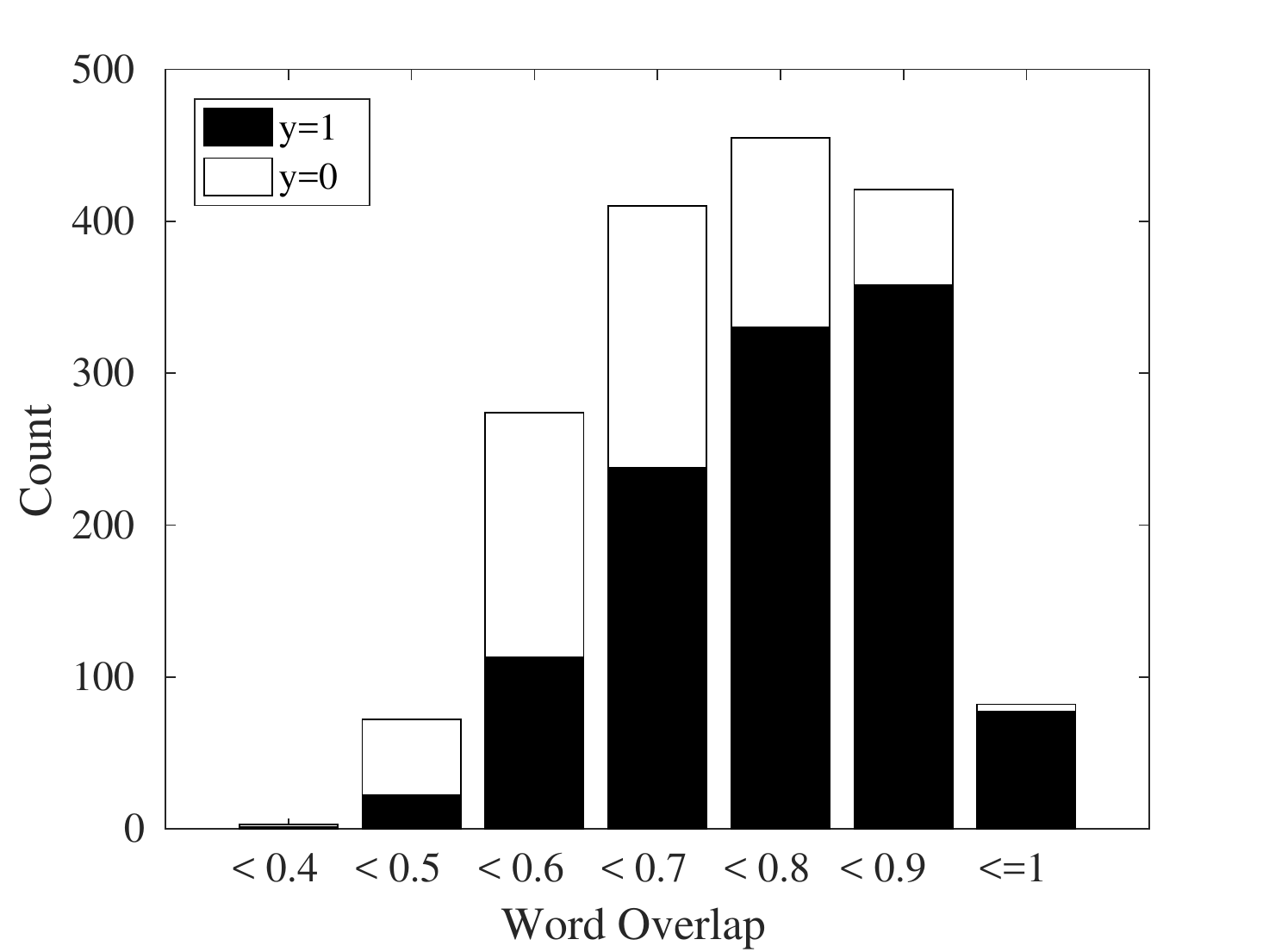}
       \captionof{figure}{Ratio of paraphrases with increasing word overlap in MSRP.}\label{fig:overlap}
       \vspace{13pt}
    \end{minipage}
  \end{figure}

 The low variance of the performance in the paraphrase detection task also reflects the overall correlation between word overlap and the likelihood of being a paraphrase as seen in Figure \ref{fig:overlap}; for difficult cases, as in the examples in Table \ref{tab:ex2}, the overall similarity is not a good indication of being a paraphrase. Significant improvements in this task may require more nuanced features as in \cite{ji2013discriminative}.


\subsection{Evaluation on Sentiment Analysis and Categorization}

\begin{table}[hb]
\centering
\resizebox{\columnwidth}{!}{%
  \begin{tabular}{|c|c|c|c|c|c|c?c|c|c|c|c|c|}
    \cline{3-13}
    \multicolumn{2}{l|}{} &
    \multicolumn{5}{c?}{Sentiment Analysis} &
    \multicolumn{6}{c|}{Text Categorization} \\
    \cline{3-13}
    \multicolumn{2}{l|}{} &
    \multicolumn{1}{c|}{CR} & \multicolumn{1}{c|}{mpqa} & \multicolumn{1}{c|}{RT-s} & \multicolumn{1}{c|}{subj} & \multicolumn{1}{c?}{imdb} &  \multicolumn{1}{c|}{rel} & \multicolumn{1}{c|}{spo} & \multicolumn{1}{c|}{com} & \multicolumn{1}{c|}{pol} & \multicolumn{1}{c|}{mult} & \multicolumn{1}{c|}{TREC} \\
    \hline
    \multicolumn{2}{|c|}{Binary BOW}  &  77.0/0.821 &85.9/0.756 &74.8 &89.5 &84.1 &66.2/0.710 &85.7 &78.2 &81.6 &  72.2 &\BW 89.8\\
     \hline
     \multicolumn{2}{|c|}{Unigram NBSVM}  & \G 80.5 & 85.3 & \G 78.1 & \BW 92.4 & \BW 88.3 & \BW 73.2 & \BW 93.5 & \BW 86.7 & \BW 91.9 &\BW 93.4 & \BW 89.8 \\
    \hline
     \specialrule{.2em}{.1em}{.1em} 
     \multicolumn{2}{|c|}{Paragraph Vector (doc2vec)}  & 76.6/0.831 &82.4/0.688 &\G 78.6 &89.9 &87.8 & \G68.9/0.751 &\GG 89.6 &\G 82.3 &\BW 90.6 &\GG \B 76.1 &59.6 \\
     \hline
    \multirow{3}{*}{si-skip}  & avg & \G 81.3/0.857 & \GG 87.2/0.778 &\G 78.8 &\GG 91.6 & \BW88.0 &67.1/0.759 &\GG 87.8 & \G 80.3 &  87.2 & \G 74.4 &  79.6\\
    \cline{2-13}
    &idf& \G 80.2/0.849 &86.2/0.765 &\G 78.5 &\GG 91.0 &87.0 &\G 69.1/0.755 & \GG \B 88.3 & \G 81.9 & \GG \B 89.7 & \GG \B 75.0 &70.4\\
    \cline{2-13}
    &sif& \G 80.6/0.852 &\GG86.7/0.774 &\G 78.6 &\GG 91.0 &\BW 88.2 & \G 69.1/0.765 &\GG \B 88.8 & \G 81.0 &  \GG \B 89.4 & \G 74.7 &71.8\\
    \hline
    \multirow{3}{*}{CBOW}  & avg & \G 81.6/0.859 &86.1/0.759 &\G 78.1 &\GG 91.0 &87.2 &63.6/0.745 &74.9 &78.2 &84.7 &66.6 & \GG \B 82.8\\
    \cline{2-13}
    &idf& \G 81.2/0.856 &86.0/0.761 & 77.7 &90.5 &87.3 &65.8/0.745 &78.6 &79.8 &85.6 &68.2 &77.4\\
    \cline{2-13}
    &sif& \G 80.8/0.852 &85.7/0.756 &77.8 &90.2 &87.4 &65.6/0.742 & 80.0 &79.5 &85. &68.3 &76.2\\
    \hline
    \multirow{3}{*}{GloVe}  & avg & \G 80.7/0.851 &85.4/0.745 &77.6 &\GG 91.0 &87.3 &67.1/0.748 &82.1 &78.8 &86.5 &69.3 &  79.8\\
    \cline{2-13}
    &idf& \G 80.7/0.851 &85.8/0.756 &77.8 &90.8 &87.5 &65.5/0.725 &85.2 &77.7 &87.6 &70.9 &72.4\\
    \cline{2-13}
    &sif& \G 80.6/0.850 &85.4/0.751 &77.6 &90.7 &87.4 &66.8/0.736 &85.8 &78.5 &87.5 &70.9 &72.0\\
    \hline
    \multirow{3}{*}{Random}  & avg & 70.7/0.779 &74.2/0.413 &61.9 &77.3 &74.0 &55.8/0.634 &71.4 &72.8 &69.4 &47. &70.2\\
    \cline{2-13}
    &idf& 68.6/0.763 &74.1/0.423 &61.1 &72.7 &74.2 &58.9/0.654 &74.4 &72.7 &72.9 &47.2 &57.0\\
    \cline{2-13}
    &sif& 69.2/0.770 &72.7/0.369 &61.5 &69.6 &74.5 &59.4/0.655 &73.0 &72.1 &72.6 &47.4 &54.8\\
    \hline
      \specialrule{.2em}{.1em}{.1em} 
     \multicolumn{2}{|c|}{GRAN}  &78.4/0.838 &86.6/0.769 &75.1 &88.5 &83.1 &66.0/0.753 &90.6 & 80.8 &  88.5 &\G 73.2 &60.4\\
     \hline
       \multirow{3}{1.9cm}{\centering pre-trained \small{PARAGRAM-SL999}$\dagger$}  & avg & 79.8/0.845 &\GG 87.8/0.794 &75.9 &\GG 89.6 & \BW 84.5 &65.3/0.721 &89.8 &78.9 &87.5 & 72.5 &\BW 83.2\\
    \cline{2-13}
    &idf& 79.1/0.840 &\GG 87.4/0.791 &75.8 & \GG 89.3 & 84.1 & \GG 68.2/0.737 &90.3 & 79.8 & 89.0 &\G 74.1 &74.6
\\
    \cline{2-13}
    &sif& 79.2/0.840	& \GG 87.5/0.789	& 75.8	& 88.7	& \BW 84.6	& \GG 69.0/0.747	& 89.7	& 79.9	& 88.9	& \G 73.1	& 75.0\\

     \hline
     \specialrule{.2em}{.1em}{.1em} 
           \multicolumn{2}{|c|}{skip-th}  &  80.4/0.851 &87.0/0.78 &76.4 &\BW 93.4 &81.8 &65.5/0.736 &70.4 &69.4 &81.5 &60.1 & \BW 88.2\\
   \hline
         \multirow{3}{1.8cm}{\centering  Pre-trained CBOW $\dagger$ }   & avg &  79.9/0.847 &\G 88.2/0.800 &\G 77.5 &90.5 &\G 85.6 &64.8/0.751 &86.6 &\GG 79.5 &\GG 85.9 &70.7 & 80.0\\
    \cline{2-13}
    &idf& 79.6/0.844 &\G 87.9/0.797 &\G 77.2 & 90.0 &\G 85.6 &\GG 68.4/0.766 &\GG 87.5 &\GG 80.5 &\GG 85.8 & \GG 72.6 &72.2\\
    \cline{2-13}
    &sif& 79.2/0.842	& \G 87.7/0.794	& \G 77.0	& 89.7	& \G 85.8	& \GG 67.7/0.761 &	\GG 87.8	& \GG 81.3	& \GG 86.9	& \GG 72.9	& 74.2 \\
    \hline  
         \specialrule{.2em}{.1em}{.1em} 
           \multicolumn{2}{|c|}{inferSent}  &  \BW 83.0/0.867 &	88.5/0.811	 & 77.1	& 91.0 &	\GG 86.4	& 68.5/0.731	& 88.6	& 80.9	& 85.2	& 74.4	& \BW 88.6\\
   \hline
         \multirow{3}{1.8cm}{\centering  Pre-trained GloVe $\dagger$ }   & avg & 80.6/0.851	& 87.9/0.793 & 	77.1	& 90.9	& 85.7	& 69.5/0.759 &	\G 90.7	& \GG 83.8	& \GG 88.3	& 75.8	& 82.2 \\
    \cline{2-13}
    &idf& 79.8/0.844	& 87.4/0.788 &	77.2	& 90.1	& 85.5	& 69.7/0.757	& \G 89.8	& \GG 83.0 &	\GG 88.8	& \G 76.9	& 75.4 \\
    \cline{2-13}
    &sif&  79.6/0.842	& 87.2/0.785 &	77.5	& 90.0 & 	85.5	& 67.8/0.742	& \G 89.7	& \GG 83.3	& \GG88.7 &	\G 76.7	& 77.0\\
    \hline  
  \end{tabular}%
}
\caption{Accuracy \% or acuracy/F1 (for unbalanced datasets) on sentiment and topic categorization tasks.  Results are shaded according to their statistical significance using a two-tailed significance test with $\alpha = 0.05$. \small $\dagger$ pre-trained word embeddings used in the model above}\label{table:sent}
\end{table}  

Table \ref{table:sent} shows the performance in sentiment analysis and  categorization tasks. Unlike in pair-wise similarity, random vectors underperformed the NBSVM and distributed models by a large margin. This underscores the importance of global and distributed features in these tasks. 
\texttt{si-skip} outperformed other word embedding models, but we observe no advantage for weighted vs. unweighted averaging. In TREC question classification and the subjectivity benchmarks, \texttt{avg} performed significantly better than both \texttt{idf} and \texttt{sif} weighted averaging. \texttt{skip-th} vectors also significantly outperformed the pre-trained vectors in these two tasks, and underperformed in all others. We surmise that the syntactic features conveyed in frequent function words and the overall structure encoded by the LSTM network in \texttt{skip-th} may provide useful clues for these two classification tasks. 
\texttt{inferSent} achieved the highest accuracy in CR sentiment task, and on par with \texttt{skip-th} and NBSVM in TREC, and it slightly underperformed the averaging models in Newsgroup categorization. In the next section, we analyze the Newsgroup and TREC datasets to shed light on intrinsic characteristics that may explain some of the performance variance.

\section{Qualitative Analysis}

\begin{figure}
\centering
\scalebox{.8}{
\begin{minipage}[b]{\linewidth}
\begin{minipage}[b]{0.17\linewidth}
\includegraphics[height=\linewidth,width=\linewidth]{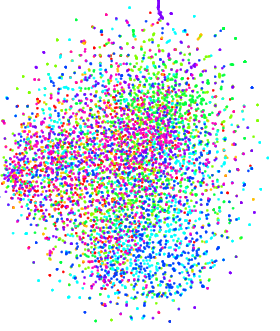}
\captionof{subfigure}{Random} \label{fig:b}
\end{minipage}
\begin{minipage}[b]{0.20\linewidth}
\includegraphics[width=\linewidth]{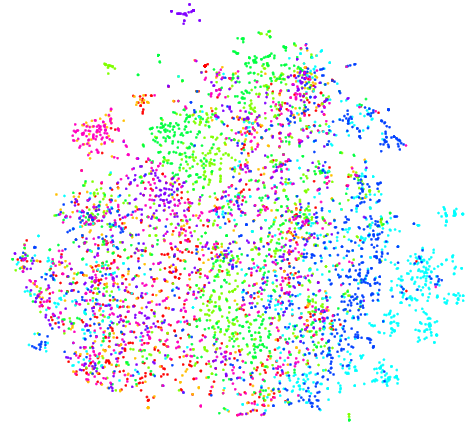}
\captionof{subfigure}{Binary} \label{fig:b}
\end{minipage}
\begin{minipage}[b]{0.20\linewidth}
\includegraphics[height=0.75\linewidth,width=\linewidth]{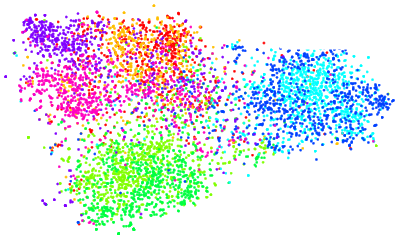}
\captionof{subfigure}{avg $\dagger$} \label{fig:b}
\end{minipage}
\begin{minipage}[b]{0.20\linewidth}
\includegraphics[width=\linewidth]{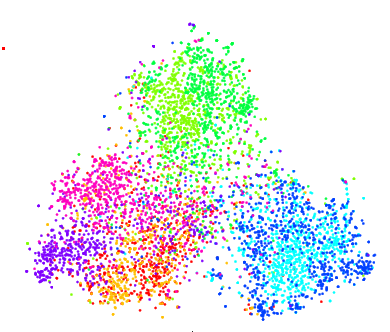}
\captionof{subfigure}{sif $\dagger$} \label{fig:b}
\end{minipage}\\
\hspace{-15pt}
\begin{minipage}[b]{0.20\linewidth}
\includegraphics[height=0.85\linewidth,width=\linewidth]{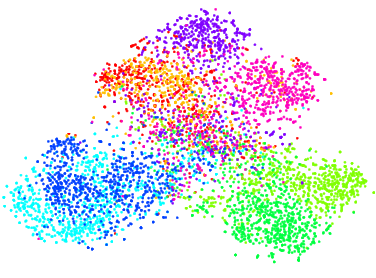}
\captionof{subfigure}{doc2vec} \label{fig:b}
\end{minipage}
\begin{minipage}[b]{0.20\linewidth}
\includegraphics[width=\linewidth]{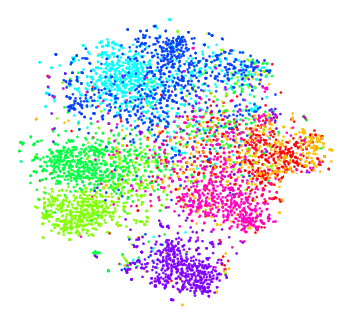}
\captionof{subfigure}{GRAN} \label{fig:b}
\end{minipage}
\begin{minipage}[b]{0.20\linewidth}
\includegraphics[width=\linewidth]{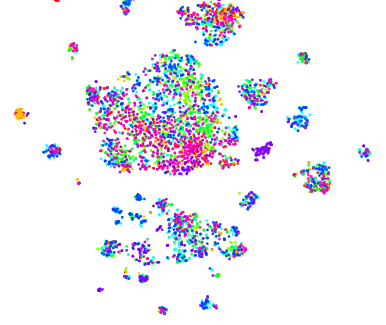}
\captionof{subfigure}{skip-th} \label{fig:b}
\end{minipage}
\begin{minipage}[b]{0.20\linewidth}
\includegraphics[width=\linewidth]{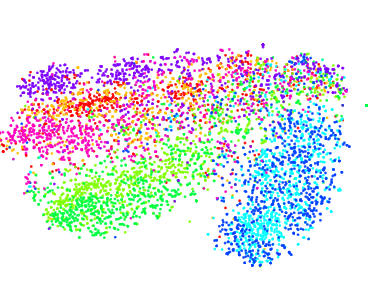}
\captionof{subfigure}{inferSent} \label{fig:b}
\end{minipage}
\begin{minipage}[b]{0.18\linewidth}
\includegraphics[width=0.6\linewidth]{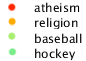} \includegraphics[width=0.6\linewidth]{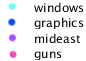}\vspace{30pt}
\end{minipage}
\end{minipage}
}
 \caption{t-SNE visualizations of  vectors on the 20-Newsgroup datasets.  {\small $\dagger$ \texttt{avg} and \texttt{sif} using \texttt{si-skip}.}} \label{fig:ng}
\end{figure}

Figure \ref{fig:ng} shows t-SNE visualizations of the Newsgroup datasets using the various compositional models, including random and binary vectors. While random and binary vectors could identify shallow similarities between sentences as in STS tasks, they failed to do so in a globally cohesive manner. The random vectors also introduced noise in the representations, which resulted in a rather uniform vector space. All other models, except \texttt{skip-th}, clearly separated at least three regions that correspond to the categories \textbf{sport}, \textbf{computer}, and \textbf{religion/politics}. Smaller clusters with consistent labeling can also be identified with minimal separation between the clusters. 

Table \ref{tab:ng_ex} shows examples of nearest neighbors using some of the models. \texttt{skip-th} vectors seem to be clustered more by structure than semantic content, unlike the \texttt{doc2vec} and \texttt{sif} models. To quantify these differences, we applied k-means clustering using $k=3$ and $k=8$, and calculated the clustering purity for each model as follows:

\begin{equation}
P(C, L)= \frac{1}{N}\sum_k \max_{\j} |c_k \cap \ell_j |
\end{equation}

\noindent where $C=\{c_1,...,c_K\}$ is the set of clusters, $L={\ell_1,...\ell_K}$ is the set of labels, and $N$ the total number of samples. As shown in Table \ref{tab:purity}, using \texttt{doc2vec} and the averaging models, including \texttt{GRAN}, k-means successfully separated the 3 categories, with \texttt{doc2vec}  and \texttt{sif} outperforming in both the fine-grained and coarse clustering. \texttt{skip-th} clusters, on the other hand, did not correspond with the correct labels, underperforming binary and random features, which explains its relatively low performance in text categorization tasks. \texttt{inferSent} achieved higher purity than binary, random and \texttt{skip-th} vectors but lower than the other models, particularly with $k=3$. 

\begin{table}[]
\small
\centering
   \scalebox{0.8}{
   \hspace{-30pt}
       \setlength{\tabcolsep}{2.5pt}
    \begin{tabular}{m{1.8cm}m{9.5cm}|m{8cm}}
        \multirow{1}{1.5cm}{\centering sif $\dagger$} & And they work especially well when the \textbf{Feds} have \textbf{cut off your utilities.} \newline  The Dividians did not have that option after the \textbf{FBI} \textbf{cut off their electricity}.  \newline Not when the \textbf{power has been cut off} for weeks on end.  & \textbf{What} does this \textbf{bill} do?   \newline  And \textbf{Bill} James is not? Do you own " the Bill James players rating book"? \newline \textbf{Who} has to consider it ? The being that does the action? I am still not sure I know what you are trying to say.\\
          \cline{2-3}
         \multirow{1}{1.5cm}{\centering doc2vec} &  And they work especially well when the \textbf{Feds} have \textbf{cut off your utilities}. \newline The Dividians did not have that option after the \textbf{FBI cut off their electricity}. \newline Can the \textbf{Feds} get him on tax evasion ? I do not remember hearing about him running to the Post Office last night. & I did not \textbf{claim} that our system was \textbf{objective}. \newline  Did I \textbf{claim} that there was an absolute morality , or just an \textbf{objective} one? \newline  I have just spent two solid months \textbf{arguing} that no such thing as an \textbf{objective} moral system exists.\\
          \cline{2-3}
         \multirow{1}{1.7cm}{\centering skip-th} & \textbf{What} does this bill do ? \newline \textbf{Where} do I get hold of these widgets ? \newline \textbf{What} gives the United States the right to keep Washington D.C.?  \newline \textbf{What} makes you think Buck will still be in New York at year 's end with George back? & \textbf{I have just spent} two solid months arguing that no such thing as an objective moral system exists. \newline  The amount of energy being spent on one lousy syllogism says volumes for the true position of reason in this group. \newline \textbf{I just heard} this week that he has started on compuserve flying models forum now.\\
      \end{tabular}
      }
      \caption{Examples of nearest neighbors in the 20-Newsgroup dataset.  {\small $\dagger$\texttt{sif} using \texttt{si-skip}.}}\label{tab:ng_ex}
\end{table}
\begin{table}[h]
\centering
    \scalebox{0.8}{
       \begin{tabular}{|c|c|c|c|}\hline
       \multirow{2}{2cm}{\centering Model}  &  \multicolumn{2}{c|}{Newsgroup} & \multicolumn{1}{c|}{TREC}\\ 
       \cline{2-4}
        & k = 3, C=3 & k = 8, C=8  & k=6, C=6\\
        \hline
        Random & 0.5563& 0.2431 & 0.4424\\
        Binary & 0.6236 & 0.2963 & 0.444 \\
        avg $\dagger$ & 0.8465 & 0.3969 & 0.4481\\
        sif $\dagger$ & \textbf{0.8776} & 0.4523 & 0.4037\\
        doc2vec & \textbf{0.8625} & \textbf{0.4967} & 0.3855\\
        skip-th & 0.4471 & 0.1854 & 0.3896\\
        GRAN & 0.8227 & 0.3553 & 0.3514\\ 
        inferSent & 0.6562 & 0.3801 & 0.4424 \\ \hline
      \end{tabular}
          }
    \captionof{table}{Clustering purity measure with coarse categories (sports, computers, religion/politics) and the original 8 categories for the Newsgroup dataset, and 6 categories for TREC. {\small $\dagger$\texttt{avg} and \texttt{sif}  using \texttt{si-skip}.} } \label{tab:purity}
\end{table}

\begin{figure}[t]
\centering
\scalebox{0.8}{
\begin{minipage}[b]{1.2\linewidth}
\centering
\begin{minipage}[b]{0.22\linewidth}
\includegraphics[width=\linewidth]{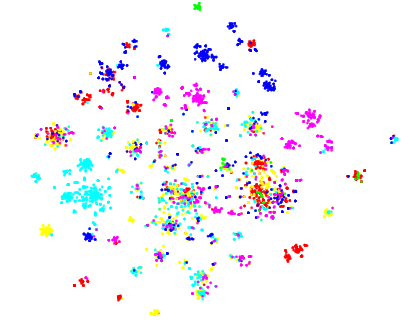}
\captionof{subfigure}{Binary} \label{fig:b}
\end{minipage}
\hspace{-10pt}
\begin{minipage}[b]{0.20\linewidth}
\includegraphics[height=0.75\linewidth,width=\linewidth]{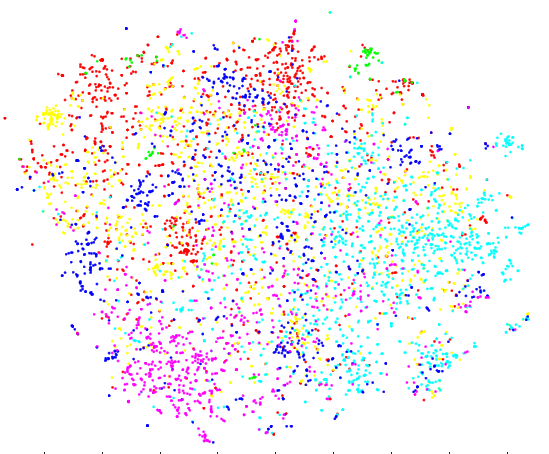}
\captionof{subfigure}{avg $\dagger$} \label{fig:b}
\end{minipage}
\begin{minipage}[b]{0.20\linewidth}
\includegraphics[width=\linewidth]{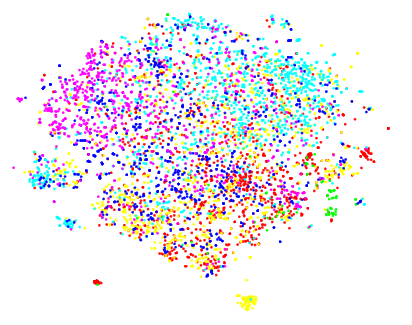}
\captionof{subfigure}{sif $\dagger$} \label{fig:b}
\end{minipage}
\hspace{-15pt}
\begin{minipage}[b]{0.20\linewidth}
\includegraphics[height=0.85\linewidth,width=\linewidth]{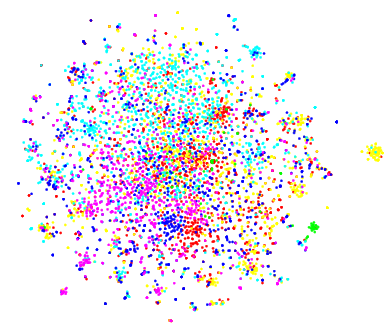}
\captionof{subfigure}{GRAN} \label{fig:b}
\end{minipage}
\\
\begin{minipage}[b]{0.12\linewidth}
\includegraphics[width=0.6\linewidth]{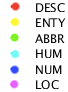} \vspace{30pt}
\end{minipage}
\hspace{-20pt}
\begin{minipage}[b]{0.18\linewidth}
\includegraphics[height=\linewidth,width=\linewidth]{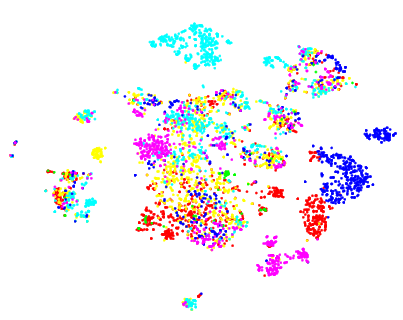}
\captionof{subfigure}{skip-th} \label{fig:b}
\end{minipage}
\begin{minipage}[b]{0.18\linewidth}
\includegraphics[height=\linewidth,width=\linewidth]{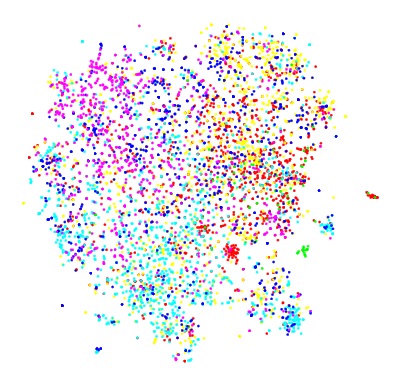}
\captionof{subfigure}{doc2vec} \label{fig:b}
\end{minipage}
\begin{minipage}[b]{0.18\linewidth}
\includegraphics[height=\linewidth,width=\linewidth]{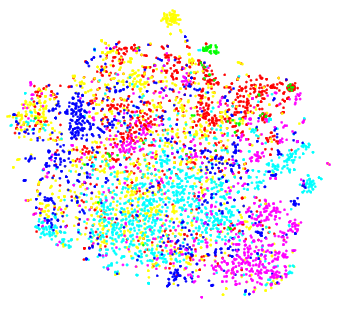}
\captionof{subfigure}{inferSent} \label{fig:b}
\end{minipage}
\end{minipage}
}

 \caption{t-SNE visualizations of  vectors on the TREC dataset. {\small $\dagger$ \texttt{avg} and \texttt{sif} using \texttt{si-skip}.}} \label{fig:trec}
\end{figure}

\begin{figure}[ht]
 \centering
   \begin{minipage}{0.58\textwidth}
    \centering
     \scalebox{0.75}{
   \setlength{\tabcolsep}{3pt}
   \hspace{-30pt}
       \begin{tabular}{m{1.5cm}m{9cm}}
        \multirow{1}{1.5cm}{\centering avg$\dagger$} &
What currents affect the area of the Shetland Islands and Orkney Islands in the North Sea? \newline
What two Caribbean countries share the island of Hispaniola?  \\
          \cline{2-2}
         \multirow{1}{1.5cm}{\centering Binary} &
What is a First World country? \newline
What is the best college in the country?  \\
          \cline{2-2}
         \multirow{1}{1.5cm}{\centering skip-th} & 
What country is the worlds leading supplier of cannabis? \newline
What country did the Nile River originate in? \newline
What country boasts the most dams?  \\
  \cline{2-2}
         \multirow{1}{1.5cm}{\centering inferSent} & 
         What country did the ancient Romans refer to as Hibernia? \newline
 How many islands does Fiji have? \newline
What country does Ileana Cotrubas come from ? \\
      \end{tabular}
      }
            \captionof{table}{Nearest neighbors to ``What country do the Galapagos Islands belong to? " in TREC. {\small $\dagger$\texttt{avg} using \texttt{si-skip}}}\label{tab:trec_ex}
  
    \end{minipage}\hfill
    \hspace{5pt}
    \begin{minipage}{.4\textwidth}
    \centering
    \includegraphics[width=1.1\textwidth]{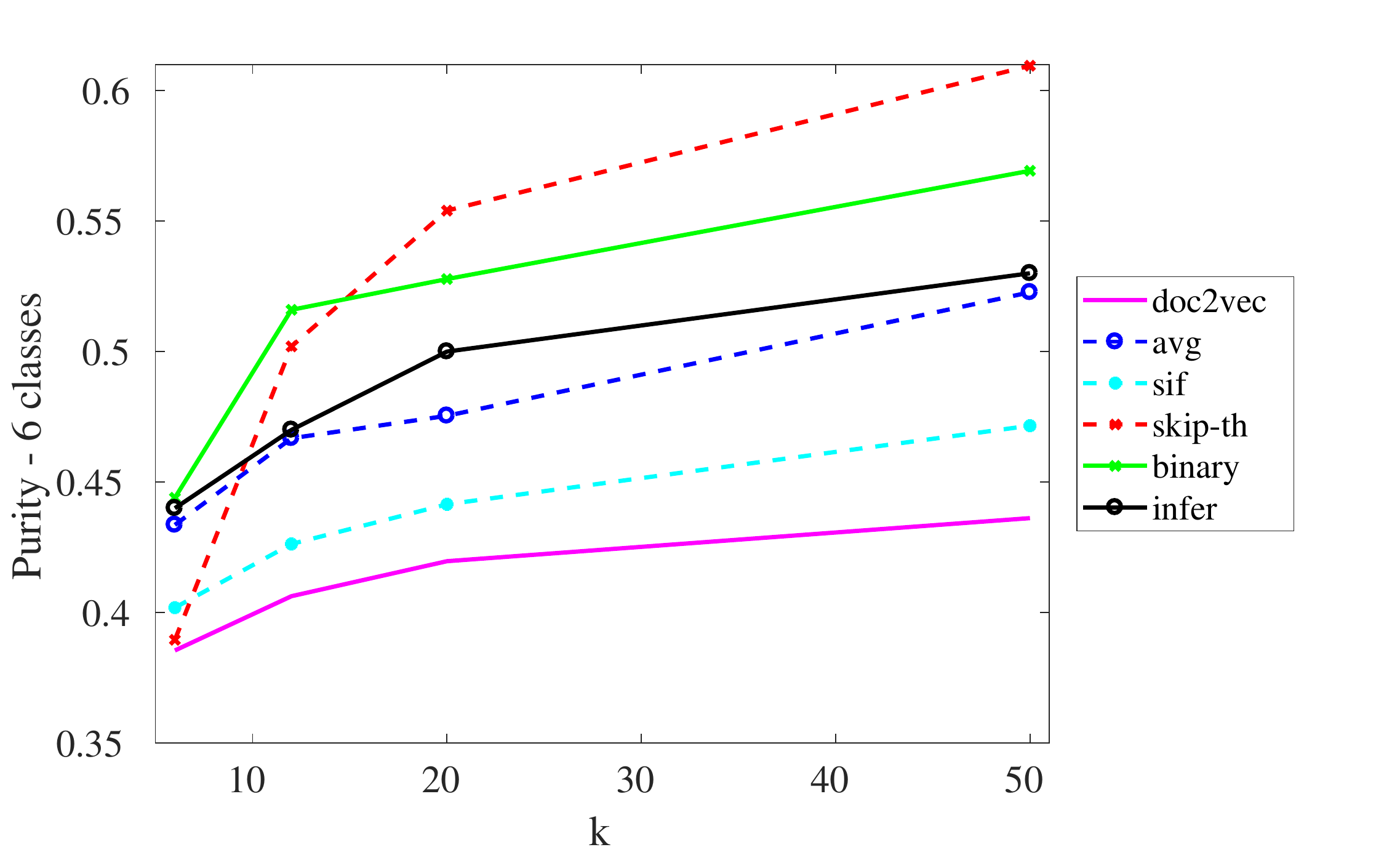}
       \captionof{figure}{Clustering purity with increasing k on TREC. {\small [\texttt{avg} and \texttt{sif}  for \texttt{si-skip}]}}\label{fig:trec_p}
    \end{minipage}
\end{figure}

Figure \ref{fig:trec} shows t-SNE visualizations of the questions in TREC training set. While all models identified some of the categories, like \textbf{HUM} and \textbf{LOC}, the \texttt{skip-th} and binary vectors appears to be more cohesively clustered by type than the other models. The question types are scattered in multiple smaller clusters, however, which explains why k-means clustering resulted in lower purity scores than \texttt{doc2vec} and \texttt{averaging} with $k=6$.  In Figure \ref{fig:trec_p}, purity results with various $k$ are plotted. While purity is expected to increase with larger $k$, the rate of increase is much higher for \texttt{skip-th} than all other models, including binary features. This is consistent with the t-SNE visualization which shows several consistent clusters with \texttt{skip-th} that are larger than the binary clusters. \texttt{inferSent}'s performance was on par with \texttt{avg}, which is slightly lower than binary and \texttt{skip-th}, although the performance in the supervised setting was equivalent. 

Table \ref{tab:trec_ex} shows nearest neighbors to the question ``What country do the Galapagos
Islands belong to?" using the various models. The averaging model clustered questions about islands; we observed similar behavior using weighted averaging, \texttt{doc2vec} and \texttt{GRAN}.  On the other hand, \texttt{skip-th} clustered questions that start with ``what country", which happens to be more suitable for identifying the \textbf{LOC} question type. Using binary vectors, questions that include the words ``What" and ``country" were clustered together, which do not necessarily correspond to the same question type.  \texttt{inferSent} vectors seem to be clustered by a combination of semantic and syntactic features.

\subsection{Discussion and Conclusions}
In this study, we attempted to identify qualitative differences among the compositional models and general characteristics of their vector spaces that explain their performance in downstream tasks.  Identifying the specific features that are most useful for each task may shed light on the type of information they encode and help optimize the representations for our needs. Word vector averaging performed reasonably well in most supervised benchmarks, and weighted averaging resulted in better performance in unsupervised similarity tasks outperforming all other models. Using the subword skipgram model for word embeddings resulted in better representations overall, particularly with \texttt{sif} weighting. The only model that performed on par with or slightly better than weighted averaging in unsupervised STS was \texttt{inferSent}.  All models achieved higher correlation scores in the supervised STS evaluation, including \texttt{skip-th}, which performed poorly in the unsupervised setting. This suggests that at least some of the features in the vector space encode the semantic content and the remaining features are superfluous or encode structural information. 

\texttt{doc2vec} and \texttt{GRAN} representations were qualitatively similar to \texttt{idf} and \texttt{sif} vectors where sentences/paragraphs were clustered by topic and semantic similarity. \texttt{skip-th} vectors, on the other hand, seemed to prominently represent structural rather than semantic features, which makes them more suitable for supervised tasks that rely on sentence structure rather than unsupervised similarity or topic categorization. \texttt{inferSent} vectors performed consistently well in all evaluation benchmarks, and a qualitative analysis of the vector space suggests that the vectors encode a balance of semantic and syntactic features. This makes \texttt{inferSent} suitable as a general-purpose model for sentence representation, particularly in supervised classification.  For topic categorization, none of the compositional models outperformed the NBSVM baseline, which achieved significantly higher accuracies in all supervised topic categorization tasks. However, the distributional models, particularly weighted averaging, are more suitable in unsupervised or low-resource settings since sentences tend to be clustered cohesively by topic similarity and semantic relatedness. 

\bibliography{refs}

\begin{thebibliography}{}

\bibitem[\protect\citename{Bojanowski \bgroup et al.\egroup
  }2017]{bojanowski2016enriching}
Piotr Bojanowski, Edouard Grave, Armand Joulin, and Tomas Mikolov.
\newblock 2017.
\newblock Enriching word vectors with subword information.
\newblock {\em Transactions of the Association of Computational Linguistics}.

\bibitem[\protect\citename{Bowman \bgroup et al.\egroup }2015]{bowmanlarge}
Samuel~R Bowman, Gabor Angeli, Christopher Potts, and Christopher~D Manning.
\newblock 2015.
\newblock A large annotated corpus for learning natural language inference.
\newblock {\em Proceedings of the 2015 Conference on Empirical Methods in
  Natural Language Processing}.

\bibitem[\protect\citename{Cer \bgroup et al.\egroup }2017]{cer2017semeval}
Daniel Cer, Mona Diab, Eneko Agirre, Inigo Lopez-Gazpio, and Lucia Specia.
\newblock 2017.
\newblock Semeval-2017 task 1: Semantic textual similarity-multilingual and
  cross-lingual focused evaluation.
\newblock {\em Proceedings of the 10th International Workshop on Semantic
  Evaluation (SemEval 2017)}.

\bibitem[\protect\citename{Chung \bgroup et al.\egroup
  }2014]{chung2014empirical}
Junyoung Chung, Caglar Gulcehre, KyungHyun Cho, and Yoshua Bengio.
\newblock 2014.
\newblock Empirical evaluation of gated recurrent neural networks on sequence
  modeling.
\newblock {\em arXiv preprint arXiv:1412.3555}.

\bibitem[\protect\citename{Conneau \bgroup et al.\egroup
  }2017]{conneau2017supervised}
Alexis Conneau, Douwe Kiela, Holger Schwenk, Lo{\"\i}c Barrault, and Antoine
  Bordes.
\newblock 2017.
\newblock Supervised learning of universal sentence representations from
  natural language inference data.
\newblock In {\em Proceedings of the 2017 Conference on Empirical Methods in
  Natural Language Processing}.

\bibitem[\protect\citename{Dolan \bgroup et al.\egroup
  }2004]{dolan2004unsupervised}
Bill Dolan, Chris Quirk, and Chris Brockett.
\newblock 2004.
\newblock Unsupervised construction of large paraphrase corpora: Exploiting
  massively parallel news sources.
\newblock In {\em Proceedings of the 20th international conference on
  Computational Linguistics}.

\bibitem[\protect\citename{Graham and Baldwin}2014]{graham2014testing}
Yvette Graham and Timothy Baldwin.
\newblock 2014.
\newblock Testing for significance of increased correlation with human
  judgment.
\newblock In {\em Proceedings of the 2014 Conference on Empirical Methods in
  Natural Language Processing (EMNLP)}, pages 172--176.

\bibitem[\protect\citename{He and McAuley}2016]{he2016ups}
Ruining He and Julian McAuley.
\newblock 2016.
\newblock Ups and downs: Modeling the visual evolution of fashion trends with
  one-class collaborative filtering.
\newblock In {\em proceedings of the 25th international conference on World
  Wide Web}.

\bibitem[\protect\citename{Hill \bgroup et al.\egroup }2016]{hill2016learning}
Felix Hill, Kyunghyun Cho, and Anna Korhonen.
\newblock 2016.
\newblock Learning distributed representations of sentences from unlabelled
  data.
\newblock pages 1367--1377.

\bibitem[\protect\citename{Hu and Liu}2004]{hu2004mining}
Minqing Hu and Bing Liu.
\newblock 2004.
\newblock Mining and summarizing customer reviews.
\newblock In {\em Proceedings of the tenth ACM SIGKDD international conference
  on Knowledge discovery and data mining}.

\bibitem[\protect\citename{Ji and Eisenstein}2013]{ji2013discriminative}
Yangfeng Ji and Jacob Eisenstein.
\newblock 2013.
\newblock Discriminative improvements to distributional sentence similarity.
\newblock In {\em Proceedings of the 2013 Conference on Empirical Methods in
  Natural Language Processing}, pages 891--896.

\bibitem[\protect\citename{Kim}2014]{kim2014convolutional}
Yoon Kim.
\newblock 2014.
\newblock Convolutional neural networks for sentence classification.
\newblock {\em Proceedings of the 2014 Conference on Empirical Methods in
  Natural Language Processing (EMNLP)}.

\bibitem[\protect\citename{Kiros \bgroup et al.\egroup }2015]{kiros2015skip}
Ryan Kiros, Yukun Zhu, Ruslan~R Salakhutdinov, Richard Zemel, Raquel Urtasun,
  Antonio Torralba, and Sanja Fidler.
\newblock 2015.
\newblock Skip-thought vectors.
\newblock In {\em Advances in neural information processing systems}, pages
  3294--3302.

\bibitem[\protect\citename{Lau and Baldwin}2016]{lau2016empirical}
Jey~Han Lau and Timothy Baldwin.
\newblock 2016.
\newblock An empirical evaluation of doc2vec with practical insights into
  document embedding generation.
\newblock {\em arXiv preprint arXiv:1607.05368}.

\bibitem[\protect\citename{Le and Mikolov}2014]{le2014distributed}
Quoc Le and Tomas Mikolov.
\newblock 2014.
\newblock Distributed representations of sentences and documents.
\newblock In {\em International Conference on Machine Learning}, pages
  1188--1196.

\bibitem[\protect\citename{Maas \bgroup et al.\egroup }2011]{maas2011learning}
Andrew~L Maas, Raymond~E Daly, Peter~T Pham, Dan Huang, Andrew~Y Ng, and
  Christopher Potts.
\newblock 2011.
\newblock Learning word vectors for sentiment analysis.
\newblock In {\em Proceedings of the 49th annual meeting of the association for
  computational linguistics: Human language technologies}.

\bibitem[\protect\citename{Marelli \bgroup et al.\egroup
  }2014]{marelli2014sick}
Marco Marelli, Stefano Menini, Marco Baroni, Luisa Bentivogli, Raffaella
  Bernardi, Roberto Zamparelli, et~al.
\newblock 2014.
\newblock A sick cure for the evaluation of compositional distributional
  semantic models.
\newblock In {\em LREC}, pages 216--223.

\bibitem[\protect\citename{Mikolov \bgroup et al.\egroup
  }2013a]{mikolov2013efficient}
Tomas Mikolov, Kai Chen, Greg Corrado, and Jeffrey Dean.
\newblock 2013a.
\newblock Efficient estimation of word representations in vector space.
\newblock {\em ICLR}.

\bibitem[\protect\citename{Mikolov \bgroup et al.\egroup
  }2013b]{mikolov2013distributed}
Tomas Mikolov, Ilya Sutskever, Kai Chen, Greg~S Corrado, and Jeff Dean.
\newblock 2013b.
\newblock Distributed representations of words and phrases and their
  compositionality.
\newblock In {\em Advances in neural information processing systems}, pages
  3111--3119.

\bibitem[\protect\citename{Milajevs \bgroup et al.\egroup
  }2014]{milajevs2014evaluating}
Dmitrijs Milajevs, Dimitri Kartsaklis, Mehrnoosh Sadrzadeh, and Matthew Purver.
\newblock 2014.
\newblock Evaluating neural word representations in tensor-based compositional
  settings.
\newblock {\em Proceedings of the Conference on Empirical Methods in Natural
  Language Processing (EMNLP)}.

\bibitem[\protect\citename{Pang and Lee}2004]{pang2004sentimental}
Bo~Pang and Lillian Lee.
\newblock 2004.
\newblock A sentimental education: Sentiment analysis using subjectivity
  summarization based on minimum cuts.
\newblock In {\em Proceedings of the 42nd annual meeting on Association for
  Computational Linguistics}, page 271.

\bibitem[\protect\citename{Pang and Lee}2005]{pang2005seeing}
Bo~Pang and Lillian Lee.
\newblock 2005.
\newblock Seeing stars: Exploiting class relationships for sentiment
  categorization with respect to rating scales.
\newblock In {\em Proceedings of the 43rd annual meeting on association for
  computational linguistics}, pages 115--124.

\bibitem[\protect\citename{Pennington \bgroup et al.\egroup
  }2014]{pennington2014glove}
Jeffrey Pennington, Richard Socher, and Christopher Manning.
\newblock 2014.
\newblock Glove: Global vectors for word representation.
\newblock In {\em Proceedings of the 2014 conference on empirical methods in
  natural language processing (EMNLP)}, pages 1532--1543.

\bibitem[\protect\citename{Riedel \bgroup et al.\egroup
  }2017]{riedel2017simple}
Benjamin Riedel, Isabelle Augenstein, Georgios~P Spithourakis, and Sebastian
  Riedel.
\newblock 2017.
\newblock A simple but tough-to-beat baseline for the fake news challenge
  stance detection task.
\newblock {\em arXiv preprint arXiv:1707.03264}.

\bibitem[\protect\citename{Schnabel \bgroup et al.\egroup
  }2015]{schnabel2015evaluation}
Tobias Schnabel, Igor Labutov, David Mimno, and Thorsten Joachims.
\newblock 2015.
\newblock Evaluation methods for unsupervised word embeddings.
\newblock In {\em Proceedings of the 2015 Conference on Empirical Methods in
  Natural Language Processing}, pages 298--307.

\bibitem[\protect\citename{Socher \bgroup et al.\egroup
  }2013]{socher2013recursive}
Richard Socher, Alex Perelygin, Jean Wu, Jason Chuang, Christopher~D Manning,
  Andrew Ng, and Christopher Potts.
\newblock 2013.
\newblock Recursive deep models for semantic compositionality over a sentiment
  treebank.
\newblock In {\em Proceedings of the 2013 conference on empirical methods in
  natural language processing}.

\bibitem[\protect\citename{Sutskever \bgroup et al.\egroup
  }2014]{sutskever2014sequence}
Ilya Sutskever, Oriol Vinyals, and Quoc~V Le.
\newblock 2014.
\newblock Sequence to sequence learning with neural networks.
\newblock In {\em Advances in neural information processing systems}, pages
  3104--3112.

\bibitem[\protect\citename{Wang and Manning}2012]{wang2012baselines}
Sida Wang and Christopher~D Manning.
\newblock 2012.
\newblock Baselines and bigrams: Simple, good sentiment and topic
  classification.
\newblock In {\em Proceedings of the 50th Annual Meeting of the Association for
  Computational Linguistics}.

\bibitem[\protect\citename{Wiebe \bgroup et al.\egroup
  }2005]{wiebe2005annotating}
Janyce Wiebe, Theresa Wilson, and Claire Cardie.
\newblock 2005.
\newblock Annotating expressions of opinions and emotions in language.
\newblock {\em Language resources and evaluation}, 39(2-3):165--210.

\bibitem[\protect\citename{Wieting and Gimpel}2017]{wieting2017revisiting}
John Wieting and Kevin Gimpel.
\newblock 2017.
\newblock Revisiting recurrent networks for paraphrastic sentence embeddings.

\bibitem[\protect\citename{Wieting \bgroup et al.\egroup
  }2015a]{wieting2015towards}
John Wieting, Mohit Bansal, Kevin Gimpel, and Karen Livescu.
\newblock 2015a.
\newblock Towards universal paraphrastic sentence embeddings.
\newblock {\em ICLR}.

\bibitem[\protect\citename{Wieting \bgroup et al.\egroup
  }2015b]{wieting2015paraphrase}
John Wieting, Mohit Bansal, Kevin Gimpel, Karen Livescu, and Dan Roth.
\newblock 2015b.
\newblock From paraphrase database to compositional paraphrase model and back.
\newblock {\em Transactions of the Association for Computational Linguistics}.

\end{thebibliography}
\bibliographystyle{acl}

\end{document}